\documentclass[10pt,journal,compsoc]{IEEEtran}
\usepackage{ragged2e}
\usepackage{color}
\usepackage{url}
\usepackage{graphicx}
\usepackage{multirow}
\usepackage{makecell}
\usepackage{array}
\usepackage{bm}
\usepackage{amsmath}
\usepackage[table]{xcolor}
\usepackage{textcomp}
\usepackage{stfloats}
\usepackage{url}
\usepackage{verbatim}
\usepackage{graphicx} 
\usepackage{algorithmic}
\usepackage{algorithm}
\usepackage{color, xcolor}
\usepackage{booktabs,array,caption}
\usepackage{multirow}
\usepackage{colortbl}
\usepackage{bbding}
\usepackage{bm}
\usepackage{enumitem}
\RequirePackage{subcaption}
\definecolor{brickred}{rgb}{0.8, 0.0, 0.0}
\definecolor{oceanblue}{RGB}{0, 0, 128}
\usepackage{makecell,tabularx}
\usepackage{amssymb}
\usepackage[colorlinks,
linkcolor=green,
anchorcolor = green,
citecolor=green]{hyperref}
\ifCLASSOPTIONcompsoc
  \usepackage[nocompress]{cite}
\else
  \usepackage{cite}
\fi

\begin{document}
%
\title{Knowledge-Informed Neural Network for Complex-Valued SAR Image Recognition}
%
%

\author{
  Haodong~Yang,
  Zhongling~Huang,~\IEEEmembership{Member,~IEEE},
  Shaojie~Guo,
  Zhe~Zhang,~\IEEEmembership{Senior~Member,~IEEE},
  Gong~Cheng,~\IEEEmembership{Senior~Member,~IEEE},
  Junwei~Han,~\IEEEmembership{Fellow,~IEEE}%
  \IEEEcompsocitemizethanks{
    \IEEEcompsocthanksitem Z. Huang, H. Yang, S. Guo, G. Cheng, and J. Han are with the School of Automation, Northwestern Polytechnical University, Xi'an, China. Z. Huang is also with the Shenzhen Research Institute of Northwestern Polytechnical University, Shenzhen, China. J. Han is also with the School of Artificial Intelligence, Chongqing University of Posts and Telecommunications, Chongqing, China.
    \IEEEcompsocthanksitem Z. Zhang is with the Aerospace Information Technology University, Jinan, China; the Suzhou Aerospace Information Research Institute, Suzhou, China; the National Key Laboratory of Microwave Imaging, Beijing, China; the Aerospace Information Research Institute, CAS, Beijing, China; and the School of Electronic, Electrical and Communication Engineering, University of Chinese Academy of Sciences, Beijing, China.
    \IEEEcompsocthanksitem The corresponding author is Zhongling Huang (huangzhongling@nwpu.edu.cn).
    \IEEEcompsocthanksitem This work was supported by the National Natural Science Foundation of China (Grant 62571443), the Guangdong Basic and Applied Basic Research Foundation (2025A1515011368), and the Natural Science Basic Research Program of Shaanxi (2025JC-QYXQ-032).
  }%
}

\markboth{Journal of \LaTeX\ Class Files,~Vol.~14, No.~8, August~2015}%
{Shell \MakeLowercase{\textit{et al.}}: Bare Demo of IEEEtran.cls for Computer Society Journals}

\IEEEtitleabstractindextext{%
\begin{abstract}
\justifying
Deep learning models for complex-valued Synthetic Aperture Radar (CV-SAR) image recognition are fundamentally constrained by a representation trilemma under data-limited and domain-shift scenarios: the concurrent, yet conflicting, optimization of generalization, interpretability, and efficiency. Our work is motivated by the premise that the rich electromagnetic scattering features inherent in CV-SAR data hold the key to resolving this trilemma, yet they are insufficiently harnessed by conventional data-driven models. To this end, we introduce the Knowledge-Informed Neural Network (KINN), a lightweight framework built upon a novel "compression-aggregation-compression" architecture. The first stage performs a physics-guided compression, wherein a novel dictionary processor adaptively embeds physical priors, enabling a compact unfolding network to efficiently extract sparse, physically-grounded signatures. A subsequent aggregation module enriches these representations, followed by a final semantic compression stage that utilizes a compact classification head with self-distillation to learn maximally task-relevant and discriminative embeddings. We instantiate KINN in both CNN (0.7M) and Vision Transformer (0.95M) variants. Extensive evaluations on five SAR benchmarks confirm that KINN establishes a state-of-the-art in parameter-efficient recognition, offering exceptional generalization in data-scarce and out-of-distribution scenarios and tangible interpretability, thereby providing an effective solution to the representation trilemma and offering a new path for trustworthy AI in SAR image analysis.

\end{abstract}

}

\maketitle


\section{Introduction}


\IEEEPARstart{D}{eep} learning has shown remarkable success in scientific and engineering domains due to its ability to learn hierarchical representations from large-scale datasets~\cite{BridgeNet, chen2025yolo, chen2024frequency, feng2024hyper, lu2025dhvt,wang2023crossformer++,ye2024invpt++}. This performance often relies on overparameterized models with high computational demands and limited interpretability, leading to a fundamental representation trilemma: balancing generalization, efficiency, and interpretability remains a major challenge in practical applications.

This issue is particularly pronounced in applications for Synthetic Aperture Radar (SAR) image, where models require generalizing well under data-limited and domain-shift scenarios, supporting lightweight deployment, and offering physical interpretability~\cite{PIHA,datcu2023explainable}. SAR is an active imaging system that produces complex-valued (CV) data containing both amplitude and phase information, capturing intricate electromagnetic scattering properties. The special data format of CV-SAR imagery makes learning an effective representation a particularly challenging task in practical scenarios~\cite{asiyabi2023complex}. Specifically, this challenge is threefold: 1) limited labeled samples and large distribution shifts hinder generalization~\cite{MFJA}; 2) compact models often lack the capacity to capture complex electromagnetic features~\cite{MS-CVNets}; and 3) insufficient interpretability undermines trust in real-world applications~\cite{ref41, MS-CVNets, PIHA}. To this end, this paper aims at CV-SAR image recognition to develop a lightweight and interpretable model with generalized representations under data-scarcity and domain-drift scenarios.

Some literature proposed to exploit the full potential of CV-SAR data using complex-valued neural networks (CVNNs)~\cite{MS-CVNets, ref41, asiyabi2023complex} to jointly process real and imaginary components. Although they better preserve electromagnetic scattering features for CV-SAR compared to amplitude-only methods, they typically require twice the parameters of real-valued networks and depend heavily on large-scale training datasets~\cite{MS-CVNets,10364955}—unrealistic in many SAR applications. Furthermore, CVNNs offer limited insight into the role of phase information, restricting their adoption in high-stakes scenarios~\cite{ref41, asiyabi2023complex}. Such opacity reduces their trustworthiness in real-world, high-stakes applications. 
Alternatively, physics-aware methods aim to enhance interpretability and reduce the dependency on large datasets by fusing pre-extracted electromagnetic priors with deep features~\cite{SDF-Net, PIHA, PAN}. However, the reliance on resource-intensive feature extraction and fusion modules introduces substantial computational overhead while also compromising the intended interpretability via opaque sub-networks and empirically-tuned hyperparameters~\cite{PIHA, PAN,CA-MCNN}. Moreover, this sensitivity to parameter settings often limits their generalization and viability for real-time applications~\cite{PIHA, FEC}.


Recent advances have explored integrating scientific priors into deep learning to improve interpretability and generalization. Some approaches embed physical models as architectural modules~\cite{KINN34} or use physics-guided loss functions~\cite{KINN12} to align outputs with known laws. Others apply information-theoretic principles, such as the Information Bottleneck, to promote compact, task-relevant representations~\cite{patel2024learning, cheng2018evaluating,gabrie2018entropy,shwartz2024compress}. While effective in natural image tasks, these methods are rarely applied to CV-SAR due to domain-specific complexities. Nonetheless, their underlying philosophies provide valuable guidance for CV-SAR model design.

As indicated in literature~\cite{shwartz2017opening} that training a deep neural network involves learning compact representations by filtering out irrelevant information while preserving task-relevant features. Models that generalize well often map data onto low-dimensional, semantically meaningful manifolds. This motivates us to develop a new framework for CV-SAR image classification that can extract informative representations with minimal parameters and limited training data. Achieving such efficiency and generalization requires effective input compression. Unlike generic visual tasks, SAR imaging is grounded in well-understood electromagnetic principles. In expert cognition, SAR target recognition relies primarily on electromagnetic scattering characteristics, rather than on background clutter or speckle noise~\cite{wang2025attributed,yifan2025asc}. This inspires us to integrate domain knowledge into the learning process, guiding the model to extract compact and physically meaningful representations from CV-SAR data under data-scarcity and domain-drift scenarios.

To achieve this, we propose the Knowledge-Informed Neural Network (KINN) for CV-SAR image recognition, and embeds scientific priors into a lightweight architecture to learn compact, physically meaningful representations. KINN is built upon a novel \textbf{“compression-aggregation-compression”} paradigm that systematically discards irrelevant information across electromagnetic and semantic spaces. Concretely, KINN integrates domain knowledge in three stages:

\textbf{1) Physics-Guided Compression:} Drawing on the Electromagnetic Scattering Center (ESC) model, a lightweight complex-valued network extracts multi-level electromagnetic representations that capture essential structural and geometric features with minimal parameters. We propose a physics-aware dictionary fine-tuning strategy based on SAR acquisition parameters to improve generalization.

\textbf{2) Adaptive Aggregation:} These electromagnetic features are transformed into the image domain and fused through a lightweight aggregation module that dynamically emphasizes informative components across levels. 

\textbf{3) Semantic Compression via Self-Distillation:} A compact classification head, equipped with a block-wise self-distillation mechanism, aligns intermediate features with low-dimensional, label-aware soft logits, enhancing discrimination while maintaining efficiency.

\begin{figure}[htbp!]
    \centering
    \includegraphics[width=0.5\textwidth]{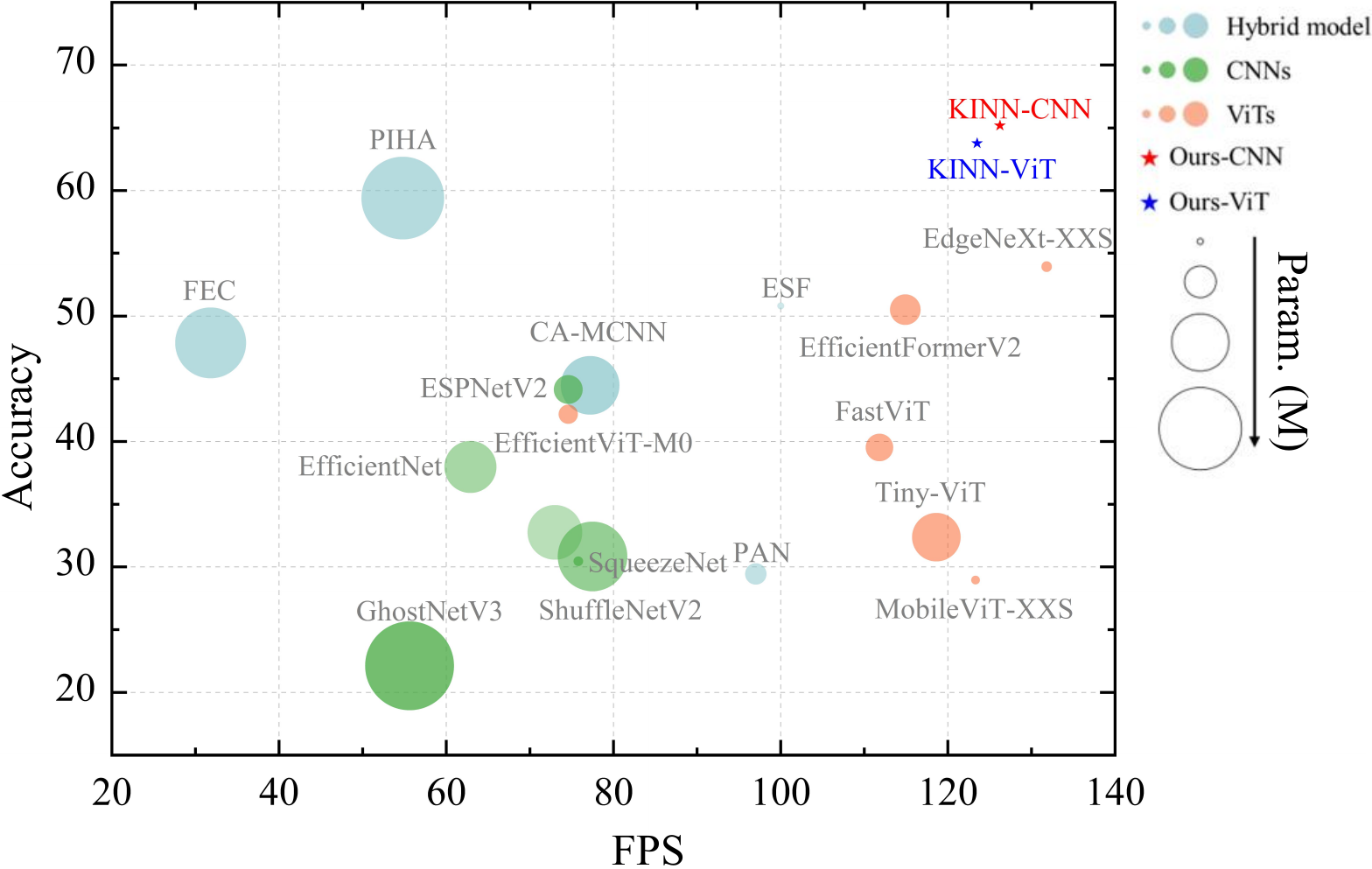}
    \caption{Comparison of KINN with state-of-the-art methods, including lightweight CNNs (green), ViTs (orange), and hybrid models for SAR (blue). KINN achieves a better performance-efficiency trade-off. All models are evaluated on the MSTAR dataset under the challenging OFA-3 protocol, using only 10\% of the available training data.}
    \label{fig:vertical_images}
\end{figure}

We instantiate this design in both CNN and Vision Transformer variants---KINN-CNN and KINN-ViT---with only 0.7M and 0.95M parameters, respectively. As shown in Fig.~\ref{fig:vertical_images}, KINN outperforms competing methods, especially under challenging generalization scenarios. Extensive experiments confirm that KINN strikes a favorable balance among generalization, interpretability, and model efficiency, while visual and quantitative analyses illustrate how KINN progressively encodes compact, task-relevant information across its layers.

The main contributions are summarized as follows: 

\begin{itemize}[leftmargin=1.5em]
    \item We propose a trustworthy knowledge-informed neural network (KINN) tailored for CV-SAR image interpretation, featuring a \textit{compression-aggregation-compression} paradigm. It achieves compact, generalizable, and interpretable representation for CV-SAR image recognition with a parameter-efficient model design.
    \item We propose a physics-inspired module grounded in electromagnetic scattering characteristics of SAR to realize interpretable and efficient compression in electromagnetic domain. The included physical parameter embedding mechanism enables generalization on various SAR imaging geometries.
    \item We thoroughly analyze how irrelevant information of CV-SAR image is effectively discarded during KINN's training, providing a mechanistic understanding of its interpretable learning paradigm.
    \item Extensive experiments on five benchmark datasets demonstrate KINN achieves new state-of-the-art performance with fewer model parameters, compared with other lightweight models as well as SAR-specific models. Notably, KINN highlights its superior generalization on data-scarcity and out-of-distribution scenarios.
\end{itemize}

\section{Related Work}
\label{sec:Rel}

\subsection{Complex-valued (CV) SAR Image Recognition}

CV-SAR imagery preserves complete electromagnetic scattering information of the observed scene, providing rich cues for object recognition. Existing methods for CV-SAR image recognition can be broadly classified into \textit{data-driven} and \textit{physics-aware} approaches. Data-driven methods primarily utilize end-to-end CVNNs to jointly process the real and imaginary components of SAR data. Several specialized architectures have been proposed, including multi-stream networks~\cite{MS-CVNets,10364955,10960712} and fully convolutional pipelines~\cite{ref41}. Other approaches improve generalization by incorporating phase-guided feature encoding strategies~\cite{ref43, ref45, ref46, ref47}. In contrast, physics-aware methods typically adopt a two-stage paradigm: electromagnetic priors—such as scattering center models—are first extracted from the CV-SAR data, and then fused with deep features learned from visual representations. These electromagnetic characteristics are encoded in diverse formats, including Bag-of-Words representations~\cite{FEC}, point cloud structures~\cite{SDF-Net, EMI-Net}, and reconstructed image components~\cite{ESF, PIHA, PAN, yifan2025asc}. 

Data-driven CVNNs offer the advantage of end-to-end trainability but typically require doubled model parameters and large-scale training datasets. Moreover, they often lack interpretability, particularly in understanding the role and contribution of phase information. In contrast, physics-aware approaches enhance interpretability by incorporating electromagnetic models, yet they tend to introduce substantial computational overhead due to the reliance on resource-intensive electromagnetic feature extraction and additional fusion modules. As a result, achieving both interpretability and efficiency for CV-SAR image recognition---while maintaining good generalization---remains a challenging trade-off in the current literature. To address this, we propose a novel architecture KINN, that balances these objectives: it employs a lightweight design that preserves physical interpretability of SAR and demonstrates strong generalization under data-scarcity and domain-drift conditions.


\subsection{Explainable Deep Models}
\label{sec:RelC}

Recent advances have explored leveraging scientific knowledge to design more explainable deep models with improved interpretability and generalization. One research direction focuses on integrating physical principles into neural network architectures~\cite{KINN34, KINN38} or introducing auxiliary loss functions grounded in physical laws~\cite{KINN12, KINN13}. These methods ensure that either the internal representations or the outputs of deep neural networks (DNNs) remain consistent with established scientific knowledge. Another line of research employs information theory to understand~\cite{shwartz2017opening} and enhance the behavior of DNNs~\cite{shwartz2024compress}. In particular, the information bottleneck principle has gained attention for its ability to explain how DNNs compress input information and to provide insights into learning compact, task-relevant representations that improve generalization~\cite{patel2024learning, cheng2018evaluating, saxe2019information, gabrie2018entropy}. Despite the success, many information bottleneck-based approaches neglect the issues of model complexity and parameter efficiency. Recently, several studies~\cite{chan2022redunet, yang2024scaling, yu2023white} have shown that incorporating information-theoretic priors into the design of white-box networks enables the learning of compact, class-discriminative representations with significantly fewer parameters, offering a promising path toward efficient and interpretable models.

Although these methods offer valuable insights, they are primarily developed for natural images and may not be fully applicable to complex-valued SAR data. In this work, we propose KINN, a model that inherits the core principles of information bottleneck theory for representation learning. Unlike existing explainable models designed for natural images, KINN is tailored to the characteristics of CV-SAR data. It learns task-relevant, compact representations by operating across both the complex-valued electromagnetic domain and the real-valued feature embedding space, following a novel “compression–aggregation–compression” paradigm.

\subsection{Electromagnetic Scattering Center Extraction Methods}
\label{sec:RelA}
The ESCs capture the dominant structural and geometric characteristics of SAR targets, yielding physically grounded representations that facilitates target recognition and interpretation.
A predominant paradigm for the extraction of ESCs over the past decades has been optimization-based approaches, motivated by the sparsity of radar echoes in the scattering center parameter space. A variety of optimization-based methods typically rely on iterative solvers, such as Orthogonal Matching Pursuit (OMP)~\cite{OMP_2, ori_omp} and Iterative Half-Thresholding (IHT)~\cite{IHT1, IHT2}. In addition, alternative frameworks have been explored, including sparse Bayesian learning~\cite{sparse_bayes, BCS}, group sparse representation~\cite{CDA}, and dictionary refinement techniques~\cite{dic_ref}. In recent years, deep learning based approaches have been developed to enable learning-based inference of scattering parameters. Notable examples include EMI-Net~\cite{EMI-Net}, which integrates sparse coding into a trainable AMP-Net, and reinforcement learning frameworks that guide scattering center extraction in an interpretable manner~\cite{reinforce}.

Despite their physical grounding, a significant challenge in optimization-based approaches is the widespread reliance on empirically-tuned hyperparameters and thresholds, which are difficult to optimize and limit generalizability. Moreover, iterative solvers within this paradigm impose a prohibitive computational burden and exhibit slow convergence, limiting their deployment in real-time applications. Conversely, deep learning paradigms offer notable improvements in computational efficiency but often entail substantial model complexity, a serious reliance on annotated data, and diminished physical interpretability. To reconcile this conflict, we propose a novel approach for ESC extraction that preserves both computational efficiency and physical interpretability, while significantly reducing model complexity and annotation requirements.

\section{Method}
\label{sec:Met}
\subsection{Preliminary}
\label{sec:ESC}

The ESC model, derived from diffraction and optics theories \cite{SC}, represents the radar echo $\bm{E(f,\varphi)}$ (dependent on frequency $f$ and aspect angle $\varphi$) as a superposition of $K_0$ individual scattering centers:
\begin{equation}
\label{equ:SC}
    \bm{E}(f,\varphi) = \sum_{i=1}^{K_0} A_i \cdot \exp\left(-j\frac{4\pi f}{c}(x_i \cos\varphi + y_i \sin\varphi)\right).
\end{equation}
Here, $A_i$ is the complex amplitude and $(x_i, y_i)$ specifies the spatial positions for the $i$-th scattering center. $j = \sqrt{-1}$ indicates the imaginary unit and $c$ is the propagation velocity of electromagnetic waves. $\bm{f}$ and $\bm{\varphi}$ represent the vectors of sampled frequency and aspect angle, respectively. 

Expanding upon the aforementioned ESC concept, the predominant portion of the radar echo $\bm{E(f,\varphi)}$ energy is derived from $K_0$ scattering centers, signifying that the echo demonstrates sparsity within the scattering center parameter space. Accordingly, the model facilitates sparse signal representation. 
Specifically, the radar echo can be vectorized into $\widetilde{s} \in \mathbb{C}^{(N_f N_\varphi)\times 1}$, where $N_f$ and $N_\varphi$ are the sampling numbers of discrete frequency and angle. The echo is then expressed as:
\begin{equation}
\label{equ:sparse_rep}
    \widetilde{s} = \bm{\widetilde{\Phi}(x,y)} \mathbf{z},
\end{equation}
where $\mathbf{z} \in \mathbb{C}^{(N_x N_y)\times 1}$ is a sparse coefficient vector encoding the complex amplitudes of scattering centers across a spatial grid defined by $(x_m, y_n)$, with $m = 1,\dots,N_x$, $n = 1,\dots,N_y$. Each element $z_{mn}$ corresponds to the complex amplitude at location $(x_m, y_n)$. 
The matrix $\bm{\widetilde{\Phi}(x,y)} \in \mathbb{C}^{(N_f N_\varphi)\times (N_x N_y)}$ is a frequency-domain dictionary whose columns are constructed based on the exponential term induced by each spatial coordinate under varying aspect angles and frequencies, following Equation~\ref{equ:SC}. 
Formally, the dictionary is expressed in column-wise form as:
\begin{equation}
\label{equ:freqD}
\begin{aligned}
    \bm{\widetilde{\Phi}(x,y)} &= [\widetilde{\Phi}_{:,1}, \widetilde{\Phi}_{:,2}, \dots, \widetilde{\Phi}_{:,N_x N_y}], \\
    \widetilde{\Phi}_{:,mn} &= \exp\left(-j\frac{4\pi \bm{f}}{c}(x_m \cos\bm{\varphi} + y_n \sin\bm{\varphi})\right).
\end{aligned}
\end{equation}
Each column of $\bm{\widetilde{\Phi}(x,y)}$ models the expected radar response from a unit-amplitude scatterer located at $(x_m, y_n)$, forming a basis for reconstructing the observed SAR echo through sparse linear combinations.

\begin{figure}[!htbp]
\centering
\includegraphics[width=0.95\linewidth]{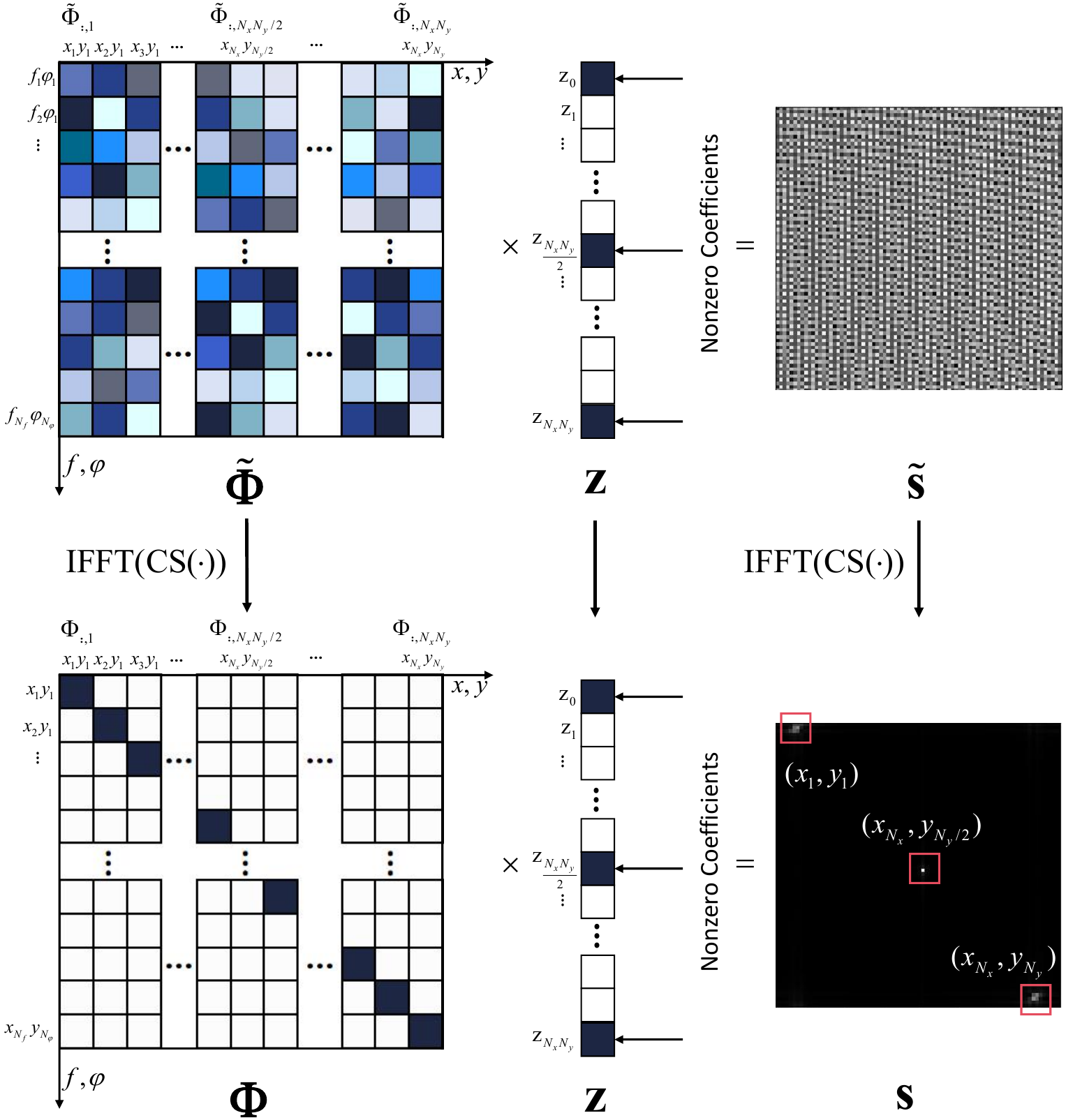}
\caption{A visualization of Equation \ref{equ:sparse_rep} in both the frequency and time domains. The upper row illustrates the frequency domain components, while the lower row depicts the image domain components subsequent to the use of Chirp Scaling (CS) algorithm and Inverse Fast Fourier Transform (IFFT).}
\label{fig:Phi}
\end{figure}

As visualized in Fig.~\ref{fig:Phi}, each column of the frequency domain dictionary encodes the frequency response associated with a specific spatial location. Notably, when transformed into the image domain, this structural sparsity allows the extraction of scattering centers in the image domain to be reframed as a sparse matching problem..

To obtain the image domain representations of both the dictionary and the signal, we apply the Chirp Scaling algorithm to compensate for range cell migration and phase distortions, followed by an inverse fast Fourier transform (IFFT):
\begin{equation}
\label{equ:timeD}
\Phi_{:,mn} = \operatorname{IFFT}\left(\text{CS}\left(\widetilde{\Phi}_{:,mn}\right)\right), \quad \mathbf{s} = \operatorname{IFFT}\left(\text{CS}\left(\mathbf{\widetilde{s}}\right)\right),
\end{equation}
where $\bm{\Phi}$ denotes the image domain dictionary composed of all transformed columns $\Phi_{:,mn}$, and $\mathbf{s}$ denotes the vectorized complex-valued SAR image.

The sparse coefficient vector $\mathbf{z}$ can then be estimated by solving the following optimization problem:
\begin{equation}
\label{equ:lasso}
    \mathbf{\hat{z}} = \mathop{\arg\min}_{\mathbf{z}} \|\mathbf{\Phi z} - \mathbf{s}\|_2^2 + \lambda\|\mathbf{z}\|_1,
\end{equation}
where $\lambda > 0$ balances data fidelity and sparsity regularization.

Regarding dimensionality, the radar echo $\bm{E}(f,\varphi)$ is obtained by discretely sampling the frequency vector $\bm{f}$ at $N_f$ points and the aspect angle vector $\bm{\varphi}$ at $N_\varphi$ points. Consequently, the vectorized form $\mathbf{\widetilde{s}}$ of the sampled echo has a dimension of $(N_f N_\varphi)\times 1$. The vectors $\bm{x}$ and $\bm{y}$ denote discrete samplings of the spatial positions, comprising $N_x$ and $N_y$ points, respectively. Thus, the sparse coefficient vector $\mathbf{z}$ has dimensions $(N_x N_y)\times 1$, and the dictionary $\bm{\widetilde{\Phi}(x,y)}$ possesses dimensions $(N_f N_\varphi)\times (N_x N_y)$. 

\subsection{KINN Overview}

A central scientific challenge in SAR image recognition lies in learning \textit{elegant} representations: compact, discriminative, and physically grounded representations derived from complex-valued SAR images. The high-dimensional and information-redundant nature of complex-valued SAR data often drives conventional models to increase network depth and parameter count in pursuit of greater expressiveness, resulting the cost of computational efficiency. The \textit{Information Bottleneck} principle offers a compelling theoretical foundation for addressing this issue by encouraging models to preserve essential information while discarding irrelevant components. Motivated by this insight, we propose a lightweight knowledge-informed neural network (KINN) that integrates SAR-specific physical priors to progressively compress and refine feature representations in a compact, interpretable, and task-aligned manner.

\begin{figure*}[!ht]
\centering
\includegraphics[width=17.78cm]{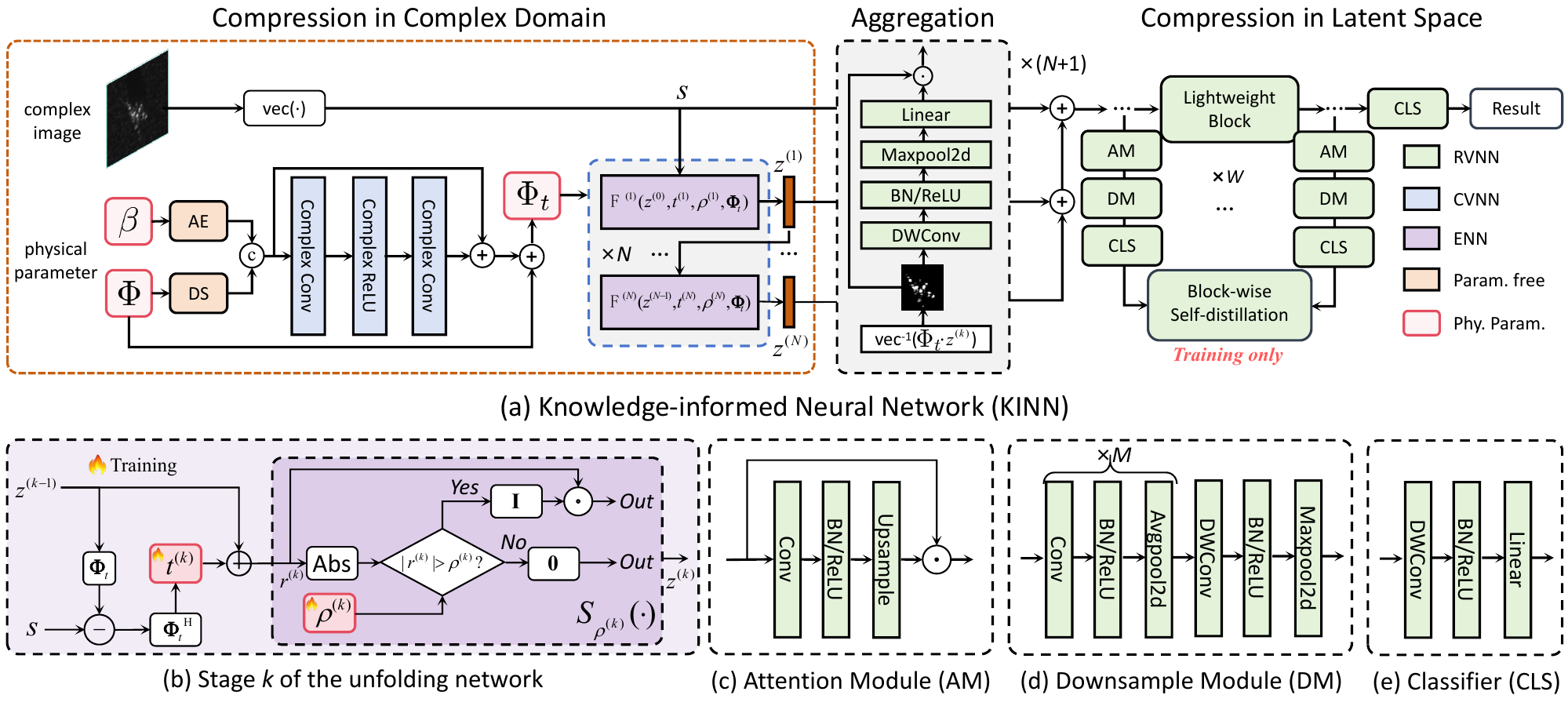}
\caption{An overview of the proposed KINN architecture. The framework (a) follows a compression-aggregation-compression paradigm. The initial compression is achieved by (b) an ISTA-based unfolding network that incorporates physical priors. Following aggregation, features are further compressed in a latent space, where each self-distillation branch composed of (c) an Attention Module, (d) a Downsampling Module, and (e) a Classifier to enhance semantic fidelity.} 
\label{fig:pipeline}
\end{figure*}

The proposed KINN follows a three-stage \textit{compression–aggregation–compression} paradigm to progressively compress and refine high-dimensional complex-valued SAR data into compact, interpretable, and task-relevant representations, as shown in Fig.~\ref{fig:pipeline}. At the initial stage, termed compression in complex domain, we aim to reduce information redundancy in complex-valued SAR images while retaining essential electromagnetic scattering characteristics. To this end, we introduce a physics-guided compression module that integrates SAR-specific priors to yield concise yet physically-grounded representations. To mitigate information loss introduced in the first stage, the resulting multi-level representations---each capturing different levels of target feature abstraction---are aggregated with the original input $\mathbf{s}$ in the second stage. Building on this enriched representation, the final stage introduces a lightweight backbone equipped with block-wise self-distillation to perform semantic-level compression with consistency across network depths. This strategy enables KINN to effectively compress and refine crucial information without relying on deep architectures, ultimately improving both interpretability and generalization.

\subsection{Compression in Complex Domain}

Given the inherent complexity of raw complex-valued SAR data, directly inputting such signals into recognition models can lead to suboptimal performance due to the presence of redundant components. To mitigate this issue, this module is designed to transform the raw data into compressive and sparse representations that retain essential target-relevant information. The module comprises two principal components. The first is a dictionary processing stage, which refines the physical dictionary $\mathbf{\Phi}$ by integrating multiple priors through the angle embedding, diagonal shear, and a lightweight complex-valued residual block, thereby producing an updated, physics-informed dictionary $\mathbf{\Phi}_t$. The second component is an ISTA-based deep unfolding network, which leverages the refined dictionary to iteratively extract sparse electromagnetic scattering center representations, yielding an interpretable and highly compressed characterization of the target’s essential electromagnetic properties. 

Specifically, the depression angle $\beta$ and the dictionary $\mathbf{\Phi}$ are first processed by the angle embedding module and the diagonal shear module, respectively, to produce structured priors $P_{\beta}$ and $P_{\mathbf{\Phi}}$:
\begin{equation}
    P_{\beta} = \text{AE}(\beta), \quad P_{\mathbf{\Phi}} = \text{DS}(\mathbf{\Phi}),
\end{equation}
where $\text{AE}(\cdot)$ and $\text{DS}(\cdot)$ represent the operations of the angle embedding module and the diagonal shear module, respectively. These outputs are concatenated and fused via the complex-valued residual block to generate the updated dictionary $\mathbf{\Phi}_t$:
\begin{equation}
    \mathbf{\Phi}_t = \text{CRB}(P_{\beta} \mathbin{\textcircled{c}} P_{\mathbf{\Phi}}) + \mathbf{\Phi},
\end{equation}
where $\text{CRB}(\cdot)$ denotes the complex residual block, and $\mathbin{\textcircled{c}}$ indicates channel-wise concatenation.

The updated dictionary $\mathbf{\Phi}_t$, along with the vectorized input $\mathbf{s}$, is subsequently fed into the deep unfolding network to obtain the sparse representation:
\begin{equation}
    \mathbf{z} = \text{DU}(\mathbf{s}, \mathbf{\Phi}_t),
\end{equation}
where $\text{DU}(\cdot)$ represents the unfolded network with $N$ learnable stages. The output $\mathbf{z}$ encodes a sparse and physically interpretable representation of the target’s ESCs. Below, we sequentially introduce the design and implementation details for each component.

\subsubsection{Angle Embedding Module}

\begin{figure}[!ht]
\centering
\includegraphics[width=5cm]{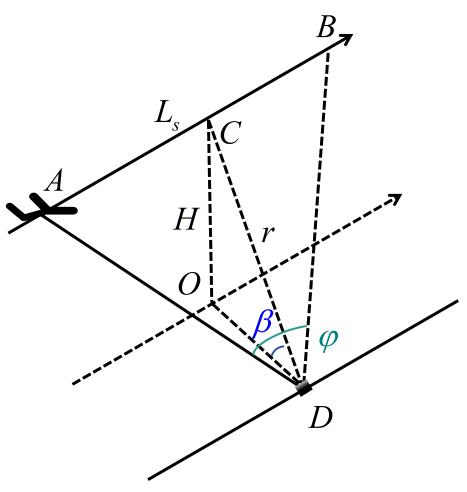}
\caption{A simplified schematic diagram of airborne radar imaging, where $\beta$ and $\varphi$ denote the depression angle and the aspect angle, respectively.} 
\label{fig:syth}
\end{figure}

\begin{figure}[H]
\centering
\includegraphics[width=\linewidth]{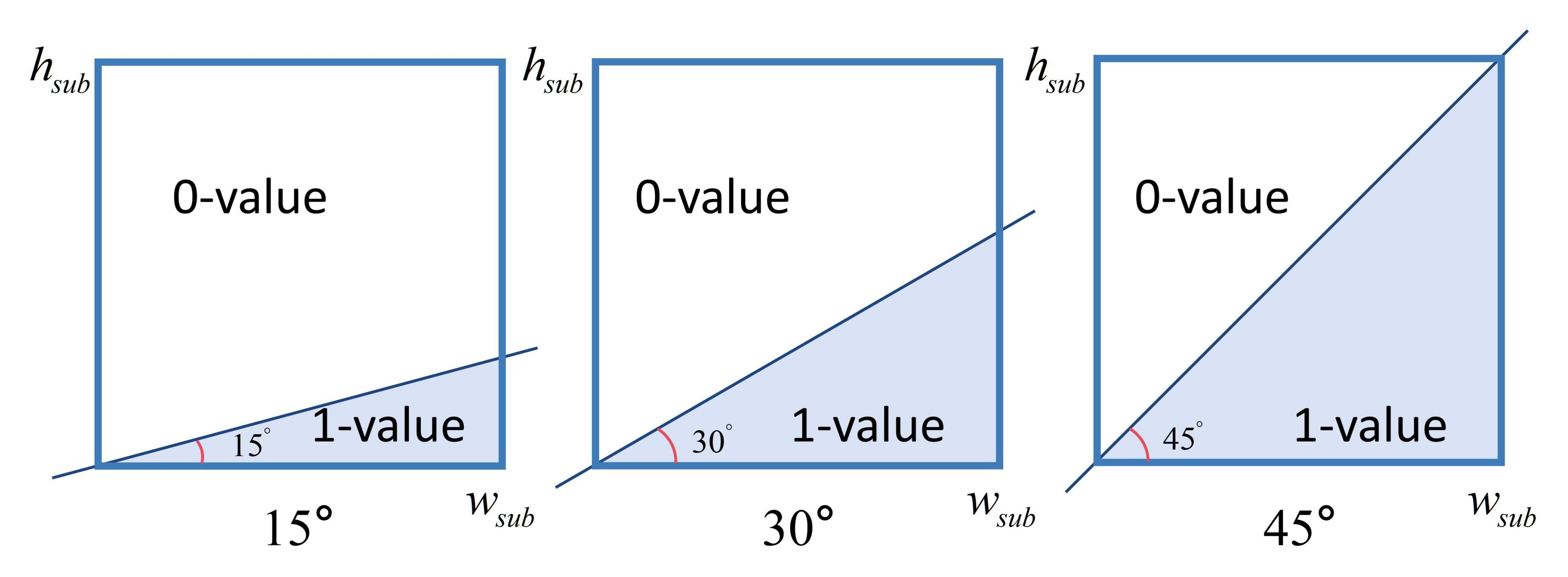}
\caption{The embedding results for depression angles of 15 degrees, 30 degrees, and 45 degrees.} 
\label{fig:Angle}
\end{figure}

Fig. \ref{fig:syth} illustrates a simplified airborne radar imaging scenario, explaining the motivation behind the angle embedding module. In this scenario, points A, B, C, O, and D represent the start point, endpoint, midpoint, nadir point corresponding to the midpoint, and the target location, respectively. $H$ and $r$ denote radar altitude and slant range, respectively, and $L_s$ indicates synthetic aperture length. $\beta$ and $\varphi$ satisfy the following geometric relation: 
\begin{equation} 
\label{equ:angle_re1} \tan\left(\frac{\varphi}{2}\right)=\frac{L_s \cdot \sin(\beta)}{2H}. 
\end{equation}
Since $H$ is typically fixed, and $L_s$ is proportional to image resolution, the aspect angle $\varphi$ correlates positively with depression angle $\beta$. To this end, we encode the depression angle $\beta$ into a structured prior matrix $P_{\beta}$ that can be integrated with $\mathbf{\Phi}$, thereby enhancing the model’s generalization capability across scenarios involving targets with varying depression angles. Specifically, as illustrated in Fig.~\ref{fig:Angle}, the embedding matrix $P_{\beta}$ is constructed as a binary matrix, where the elements located on or below a line originating from the matrix corner at an angle of $\beta$ are set to 1, and the remaining elements are set to 0.

\subsubsection{Diagonal Shear}

\begin{figure}[htbp]
\centering
\includegraphics[width=0.95\linewidth]{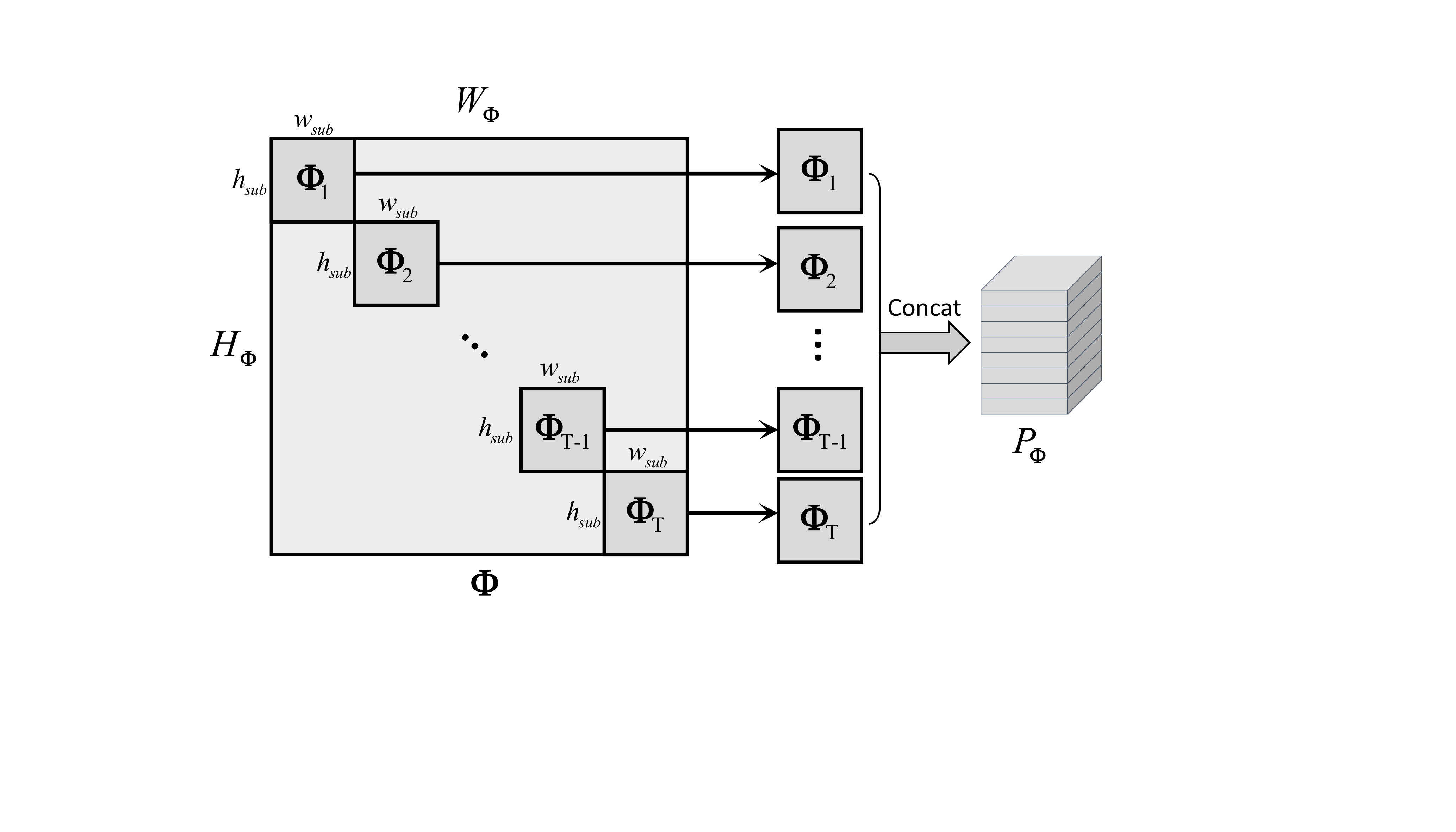}
\caption{The procedure of diagonal shear module.} 
\label{fig:DS}
\end{figure}

As previously discussed in Section \ref{sec:ESC}, the dictionary $\bm{\Phi}$ can be approximated as a sparse matrix with significant energy concentrated along the diagonal. As shown in Fig. \ref{fig:DS}, to efficiently perform shear transformation and reduce computational complexity, we partition $\bm{\Phi}$ into $T$ non-overlapping diagonal chips $\bm{\Phi}_i \in \mathbb{R}^{h_{sub} \times w_{sub}}$, $i=1,\dots, T$, where $h_{sub}$ and $w_{sub}$ are chip dimensions. The detailed process is summarized in Algorithm \ref{Diagonal Shear}, where we empirically set $T = 20$. To enable the embedding matrix $P_{\beta}$ to be concatenated with $P_{\bm{\Phi}}$, its height and width are set to match those of $P_{\bm{\Phi_i}}$.

\begin{algorithm}[!h]
    \caption{Algorithm of Diagonal Shear}
    \label{Diagonal Shear}
    \renewcommand{\algorithmicrequire}{\textbf{Input:}}
    \renewcommand{\algorithmicensure}{\textbf{Output:}}
    
    \begin{algorithmic}[1]
        \REQUIRE Dictionary $\bm{\Phi}$, Number of chips $T$
        \ENSURE Output $P_{\bm{\Phi}}$
        
        \STATE $H_{\bm{\Phi}}, W_{\bm{\Phi}} \gets \text{shape of } \bm{\Phi}$
        \STATE $h_{\text{sub}} = \left\lceil \frac{H_{\bm{\Phi}}}{T} \right\rceil$
        \STATE $w_{\text{sub}} = \left\lceil \frac{W_{\bm{\Phi}}}{T} \right\rceil$
        \STATE $P_{\bm{\Phi}} \gets \text{empty 3D array of size } T \times h_{\text{sub}} \times w_{\text{sub}}$
        
        \FOR{$i$ from $1$ \textbf{to} $T$}
            \STATE $\text{start\_row} = (i - 1) \times h_{\text{sub}}$
            \STATE $\text{end\_row} = \min(i \times h_{\text{sub}}, H_{\bm{\Phi}})$
            \STATE $\text{start\_col} = (i - 1) \times w_{\text{sub}}$
            \STATE $\text{end\_col} = \min(i \times w_{\text{sub}}, W_{\bm{\Phi}})$
            \STATE $\bm{\Phi}_i = \bm{\Phi}[\text{start\_row}:\text{end\_row}, \text{start\_col}:\text{end\_col}]$
            \STATE $P_{\bm{\Phi}}[i] = \bm{\Phi}_i$
        \ENDFOR
        
        \RETURN $P_{\bm{\Phi}}$
    \end{algorithmic}
\end{algorithm}

\subsubsection{Complex-valued Residual Block}

To efficiently integrate directional and structural priors into the physical dictionary without incurring excessive parameter costs or compromising information fidelity, a lightweight complex-valued residual block is employed. It is designed to enhance the representational capacity of the dictionary $\mathbf{\Phi}$ by fusing the outputs of the angle embedding and diagonal shear modules, ultimately producing an updated, physics-aware dictionary $\mathbf{\Phi}_t$. This block is constructed using three lightweight complex-valued convolutional layers combined with a shortcut connection, ensuring efficient prior fusion while maintaining model compactness.

\subsubsection{ISTA-based Unfolding Network}

Traditional sparse reconstruction methods for ESC extraction often suffer from fixed hyperparameters, slow convergence, and poor adaptability to data distributions. To address these limitations, we adopt a deep unfolding strategy that transforms Iterative Shrinkage-Thresholding Algorithm (ISTA) \cite{beck2009fast} into a trainable network, where each iteration is modeled as a learnable stage. Given the vectorized SAR image $\mathbf{s}$ and the refined dictionary $\bm{\Phi}_t$, the network iteratively estimates sparse coefficients, enabling the extraction of compressive and physically interpretable representations.

The traditional ISTA algorithm solves the sparse coding problem via:
\begin{equation}
\label{equ:tradISTA}
\begin{split}
    \mathbf{x}^{(k)} &= \mathbf{z}^{(k-1)} - t\bm{\Phi}_t^{H}(\bm{\Phi}_t \mathbf{z}^{(k-1)} - \mathbf{s}), \\
    \mathbf{z}^{(k)} &= S_\rho(\mathbf{x}^{(k)}),
\end{split}
\end{equation}
where $t$ is the step size, $\rho$ is the threshold, and $S_\rho(\cdot)$ denotes the soft-thresholding operator for complex inputs:
\begin{equation}
\label{equ:soft_threshold}
    S_\rho(x) = \text{sign}(x) \cdot \max(|x| - \rho, 0), \quad \text{sign}(x) = \begin{cases} \frac{x}{|x|}, & |x| > 0, \\ 0, & |x| = 0. \end{cases}
\end{equation}

To improve flexibility and efficiency, we unfold ISTA into a trainable architecture by learning stage-specific parameters $\{t^{(k)}, \rho^{(k)}\}$ at each iteration. The $k$-th stage of the unfolded network is defined as:
\begin{equation}
\label{equ:unfoldISTA}
    \mathbf{z}^{(k)} = S_{\rho^{(k)}}\left(\mathbf{z}^{(k-1)} + t^{(k)}\bm{\Phi}_t^{H}(\mathbf{s} - \bm{\Phi}_t \mathbf{z}^{(k-1)})\right).
\end{equation}

We set the number of stages $N = 3$, resulting in six learnable parameters. Unlike conventional unfolding approaches, our design explicitly incorporates SAR-specific priors into both dictionary refinement and sparse inference, enabling interpretable, adaptive, and efficient representation learning for electromagnetic scattering.

To ensure fidelity during compression, we define the reconstruction loss as:
\begin{equation} 
\label{equ:loss}
    L_c = \|\mathbf{s} - \hat{\mathbf{s}}^{(N)}\|_2^2 + \lambda\|\mathbf{z}^{(N)}\|_1, 
\end{equation} 
where $\hat{\mathbf{s}}^{(N)} = \bm{\Phi}_t \mathbf{z}^{(N)}$, and $\lambda$ is empirically set to 300.

\subsection{Aggregation}

Each phase of the proposed network yields progressively compressed representations that capture different levels of sparsity and task-relevant information. However, prior studies \cite{PIHA, FEC, PAN, CA-MCNN, EMI-Net} typically focus only on ESC parameters or final outputs, neglecting intermediate reconstructions generated during iterative extraction. This limits the ability to exploit multi-stage representations and capture complementary structural cues, thereby compromising interpretability and robustness.

To address these limitations, we introduce an aggregation module that adaptively fuses intermediate reconstructed images $\{\hat{\mathbf{s}}^{(1)}, \dots, \hat{\mathbf{s}}^{(N)}\}$ with the original input $\mathbf{s}$, where each $\hat{\mathbf{s}}^{(k)} = \mathbf{\Phi}_t \mathbf{z}^{(k)}$ is derived from the $k$-th stage of the unfolding network. As shown in Fig.~\ref{fig:pipeline}, each image is independently processed by a lightweight fusion unit composed of devectorization, depthwise convolution (DWConv), batch normalization (BN), ReLU, max pooling, and a linear projection, yielding adaptive weights $\{\gamma_{(1)}, \dots, \gamma_{(N+1)}\}$. These weights determine the relative importance of each representation in computing the aggregated output:
\[
\mathbf{s}_{F} = \gamma_{(N+1)} \cdot \mathbf{s} + \sum_{i=1}^{N}\gamma_{(i)} \cdot \mathbf{\hat{s}}^{(i)}.
\]
Here, each scalar $\gamma_{(i)}$ reflects the contribution of the corresponding representation to the final recognition, enhancing both feature diversity and model interpretability.

\subsection{Compression in Latent Space}
To obtain compressive and highly discriminative latent representations, this stage refines the aggregated features from the previous modules while ensuring consistency across network depths to enhance generalization and robustness. As illustrated in Fig.~\ref{fig:pipeline}, the final stage, termed \textit{compression in latent space}, consists of a backbone network equipped with $W$ intermediate branches inserted after early feature extraction blocks, collectively forming a dedicated self-distillation mechanism. It not only facilitates early-stage prediction and strengthens alignment between intermediate representations and final task predictions, but also achieves effective information compression by efficiently condensing critical task-relevant features without reliance on deep architectures, thereby enhancing both feature compactness and training efficiency.

\subsubsection{Backbone Network}  
To ensure architectural flexibility, we instantiate the backbone using either Convolutional Neural Networks (CNNs) or Vision Transformers (ViTs), depending on the experimental configuration. 

For CNN-based backbones, we adopt the block design from real-valued MSNet \cite{MS-CVNets}. To reduce computational complexity and enhance lightweight deployment, all standard convolutions in the MSNet blocks are replaced with DWConv. For ViT-based backbones, we employ the MobileViT \cite{MobileViT} architecture, which integrates convolutional inductive bias into transformer blocks, offering a favorable trade-off between accuracy and efficiency. 

In both cases, the backbone is composed of $W$ sequential feature extraction blocks, denoted as $f_1(\cdot), f_2(\cdot), \dots, f_W(\cdot)$, followed by a final classifier $c(\cdot)$. The detailed configurations of each backbone will be presented in the experimental section.

\subsubsection{Block-wise Self Distillation}
To achieve explicit compression within the latent space and ensure that multi-depth features contribute effectively to the final task, we incorporate a block-wise self-distillation mechanism. Specifically, an auxiliary branch is attached after each feature extraction block of the backbone. Each branch comprises three components: an attention module, a downsampling module, and a classifier. The attention module consists of a convolutional layer followed by an upsampling operation; the resulting upsampled features are element-wise multiplied with the corresponding input feature map to selectively emphasize salient regions. The downsampling module includes $M$ convolutional layers, each followed by batch normalization, ReLU activation, and average pooling, enabling a gradual reduction of redundancy and compression of feature dimensionality. Subsequently, a depthwise convolution, along with batch normalization, ReLU activation, and max pooling, is applied to further enhance spatial compactness and prepare the representations for final classification. The classifier is responsible for generating early-stage logits that serve as supervisory signals for the self-distillation process.

The logits produced by each auxiliary branch and the final classifier of the backbone are denoted as $l_1(\mathbf{s}_{F}), l_2(\mathbf{s}_{F}), \dots, l_W(\mathbf{s}_{F}), l_{W+1}(\mathbf{s}_{F})$, where $l_j(\mathbf{s}_{F})$ represents the logits from the $j$-th branch and $l_{W+1}(\mathbf{s}_{F})$ denotes the output from the final backbone classifier. A unified \textit{teacher logit} is then computed by averaging all outputs, which is expressed as $l_t(\mathbf{s}_{F}) = \frac{1}{W+1} \sum_{j=1}^{W+1} l_j(\mathbf{s}_{F})$.
Given the ground truth label $y$, the overall recognition loss is defined as:
\begin{equation}
\label{equ:latent-loss}
L_r = \frac{1}{W+1} \sum_{j=1}^{W+1} \left[ L_{CE}(l_j(\mathbf{s}_{F}), y) + L_{KL}(l_j(\mathbf{s}_{F}), l_t(\mathbf{s}_{F})) \right],
\end{equation}
\noindent where $L_{CE}$ denotes the cross-entropy loss, and $L_{KL}$ is the Kullback-Leibler divergence measuring the distance to the teacher output.

It is important to note that the self-distillation strategy is only used during training. During inference, only the backbone and its final classifier are retained, ensuring no additional computational overhead.




\section{Experiments}
\label{sec:Exp}



\subsection{Datasets and Experimental Setup}
\subsubsection{Datasets}
\noindent \textbf{MSTAR \cite{mstar}.} The MSTAR dataset, provided by Sandia National Laboratories, contains X-band SAR imagery of ten military vehicle types. The data was captured by a Twin Otter sensor at various depression angles (15\textdegree{}, 17\textdegree{}, 30\textdegree{}, and 45\textdegree{}), making it a standard benchmark for evaluating performance under different viewing conditions.

\noindent \textbf{OpenSARShip \cite{opensarship}.} The OpenSARShip dataset consists of C-band SAR ship imagery from the Sentinel-1 satellite with a 20-meter spatial resolution. We utilize its Single-Look Complex (SLC) data, which covers three main categories: Cargo vessels, Tankers, and Other. This dataset is used to evaluate model performance across targets of significantly different scales (Small, Middle, and Large). 

\noindent \textbf{CSRSDD \cite{CSRSDD}.} The CSRSDD dataset offers high-resolution (1m) ship images acquired in the GF-3 satellite's Spotlight mode. Based on the provided annotations, we cropped the ship slices to construct a recognition dataset comprising seven target types, such as aircraft carriers, amphibious ships, and destroyers.


\noindent \textbf{SAR-Aircraft-1.0 \cite{aircraft}.} The SAR-Aircraft-1.0 dataset provides 16,463 aircraft instances collected by the GF-3 satellite in a 1m-resolution Spotlight mode. It is categorized into seven classes of common aircraft, including the A220, A320/321, A330, ARJ21, Boeing787, Boeing737 and and other, serving as a key benchmark for high-resolution aircraft recognition.


\noindent \textbf{SAMPLE \cite{SAMPLE}.}The SAMPLE dataset uniquely combines real and simulated SAR targets across ten categories of military vehicles. Due to the substantial domain discrepancy between its synthetic and measured data \cite{9356129}, we utilize this dataset to rigorously assess the model's out-of-distribution (OOD) generalization capability.

\subsubsection{Experimental Setup}

\textbf{Dataset Preparation.} We evaluate the proposed KINN under two rigorous conditions: (1) limited training samples and (2) out-of-distribution (OOD) generalization. During the training of the compression in complex domain, only five training samples per category in each dataset were used. For target recognition, the MSTAR dataset \cite{mstar} is evaluated using the Once-For-All (OFA) protocol \cite{PIHA}, with training subsets (50\%, 30\%, and 10\% of the original data) and domain-variant test scenarios (OFA-2 and OFA-3). The OpenSARShip \cite{opensarship}, CSRSDD \cite{CSRSDD}, SAR-Aircraft-1.0 \cite{aircraft}, and SAMPLE \cite{SAMPLE} datasets are tested under varying training proportions and OOD conditions, as shown in Table \ref{tab:dataset}. The input SAR images are processed with L2 normalization and transformed to 80$\times$80.





\begin{table}[htbp]\footnotesize
  \centering
  \caption{The details of the training set and test set of OpenSARShip dataset, CSRSDD dataset, SAR-Aircraft-1.0 dataset and SAMPLE dataset for target recognition.}
  \resizebox{1.0\linewidth}{!}{
    \begin{tabular}{ccccc}
    \toprule
    \multirow{2}[2]{*}{Dataset} & \multirow{2}[2]{*}{Class Name} & \multirow{2}[2]{*}{Scale} & \multicolumn{2}{c}{Instance No.} \\
\cmidrule{4-5}          &       &       & Train & Test \\
    \midrule
    \multirow{3}{*}{OpenSARShip} & \multicolumn{1}{c}{\multirow{3}{*}{\makecell{Cargo, \\ Tanker, \\
    Other Type}}} & Small & 664 & 1420 \\
          &       & Medium & 1140  & 2374 \\
          &       & Large & 402 & 1026 \\
    \bottomrule
    \toprule
    \multirow{2}[2]{*}{Dataset} & \multirow{2}[2]{*}{Class Name} & \multirow{2}[2]{*}{Ratio} & \multicolumn{2}{c}{Instance No.} \\
\cmidrule{4-5}          &       &       & Train & Test \\
    \midrule
    \multirow{4}{*}{CSRSDD} & \multicolumn{1}{c}{\multirow{4}{*}
    {\makecell{Carrier, Amphibious,\\
    Cargo, Depot ship, Destroyer,\\
    Light boat, other}}} & 10\%  & 94   & \multirow{4}{*}{954} \\
          &       & 30\%  & 284  &  \\
          &       & 50\%  & 476  &  \\
          &       & 100\% & 951  &  \\
    \bottomrule
    \toprule
    \multirow{2}[2]{*}{Dataset} & \multirow{2}[2]{*}{Class Name} & \multirow{2}[2]{*}{Ratio} & \multicolumn{2}{c}{Instance No.} \\
\cmidrule{4-5}          &       &       & Train & Test \\
    \midrule
    \multirow{4}{*}{SAR-Aircraft-1.0} & \multicolumn{1}{c}{\multirow{4}{*}{\makecell{A220, A320/A321,\\
    A330, ARJ21, Boeing737,\\
    Boeing787, other}}} & 10\%  & 823   & \multirow{4}{*}{8232} \\
          &       & 30\%  & 2469  &  \\
          &       & 50\%  & 4115  &  \\
          &       & 100\% & 8231  &  \\
    \bottomrule
    \toprule
    \multirow{2}[2]{*}{Dataset} & \multirow{2}[2]{*}{Class Name} & \multirow{2}[2]{*}{Ratio} & \multicolumn{2}{c}{Instance No.} \\
\cmidrule{4-5}          &       &       & Train & Test \\
    \midrule
    \multirow{4}{*}{SAMPLE} & \multicolumn{1}{c}{\multirow{4}{*}{\makecell{2S1, BMP-2,BTR-70\\
    ZSU-234, T72, m1, m2, \\
    m35, m60, m548}}} & 10\%  & 134   & \multirow{4}{*}{1345} \\
          &       & 30\%  & 403  &  \\
          &       & 50\%  & 672  &  \\
          &       & 100\% & 1345  &  \\
    \bottomrule
    \end{tabular}}
  \label{tab:dataset}
\end{table}

\noindent \textbf{Implementation Details.} AdamW \cite{AdamW} optimizer with OneCycleLR \cite{LR} learning rate scheduler is applied. The initial learning rate is set to $2e\textnormal{-}4$, and the weight decay is 0.05. The number of training epochs and the batch size are set to 100 and 16, respectively. All experiments are conducted on a GeForce RTX 3090 GPU. 

The number of unfolding stages \textit{N} is set to 3. In each stage $k$, the initial values of $t^{(k)}$ and $\rho^{(k)}$ are set to 0.01 and 0.005, respectively. If the depression angles are not available in the dataset, the angle embedding module will be removed. The number of CNN/ViT sequential feature extraction blocks is set to 3, and the hyperparameter $M$ for downsample module in each auxiliary branch is set to 3, 3, and 1, respectively.




\subsection{Effectiveness on ESC Optimization}

\subsubsection{ESC Parameter Estimation}
\label{DU}

We evaluate the performance of the estimated ESCs through two metrics. First, we assess parameter inversion quality by comparing Peak Signal-to-Noise Ratio (PSNR) between SAR images reconstructed utilizing the estimated physical parameters by the proposed and conventional approaches, where higher PSNR indicates better inversion accuracy and reconstruction fidelity. Second, we compare the recognition performance of various SAR-ATR methods using ESC features extracted by our method versus those obtained by OMP \cite{OMP}.


\begin{figure}[H]
\centering
\includegraphics[width=1.0\linewidth]{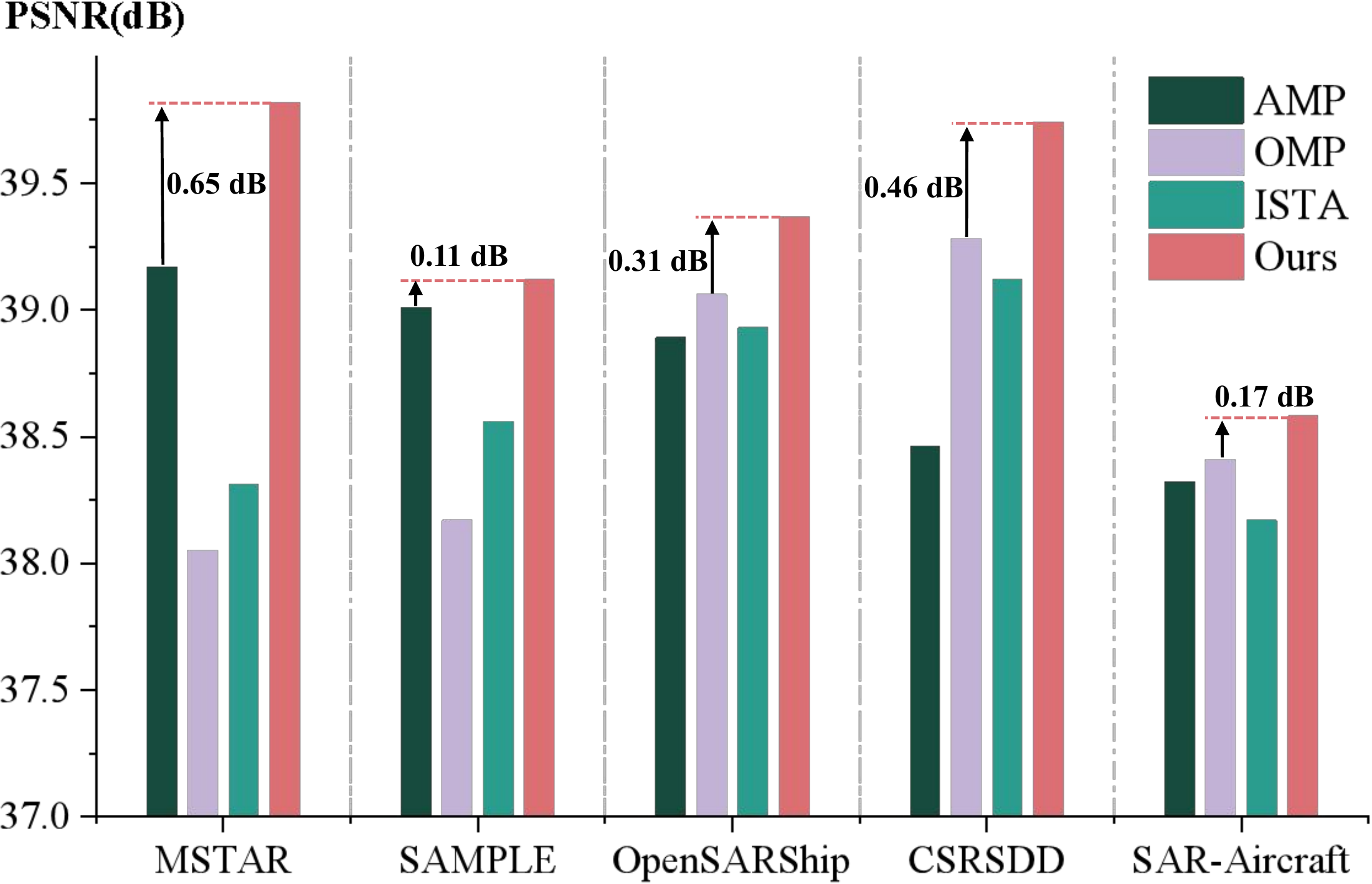}
\caption{Comparison of PSNR between traditional methods and ours.} 
\label{fig:recon_trad}
\end{figure}

\begin{table*}[!ht]
  \centering
  \setlength{\tabcolsep}{4pt}
  \caption{Performance comparison of SAR target recognition models using ESC features extracted by OMP versus our method. The experiments are conducted on MSTAR dataset with OFA evaluation protocol.}
  \resizebox{0.95\linewidth}{!}{
    \begin{tabular}{c@{\hskip 1pt}|ccc|ccc|ccc|ccc|c}
    \toprule
    \rowcolor[rgb]{ .929, .929, .929}
     &
    \multicolumn{3}{c|}{\textbf{90\%}} &
    \multicolumn{3}{c|}{\textbf{50\%}} &
    \multicolumn{3}{c|}{\textbf{30\%}} &
    \multicolumn{3}{c|}{\textbf{10\%}} &
     \\
    \rowcolor[rgb]{ .929, .929, .929}
    \multirow{-2}{*}{\textbf{Method}}
    & \textbf{OFA1} & \textbf{OFA2} & \textbf{OFA3}
    & \textbf{OFA1} & \textbf{OFA2} & \textbf{OFA3}
    & \textbf{OFA1} & \textbf{OFA2} & \textbf{OFA3}
    & \textbf{OFA1} & \textbf{OFA2} & \textbf{OFA3}
    & \multirow{-2}{*}{\textbf{Average}}\\
    \midrule
    FEC\textsuperscript{\cite{FEC}} (w/ OMP\textsuperscript{\cite{OMP}})
    & 92.32 & 86.22 & 58.76 & 86.23 & 81.34 & 55.03 & 68.43 & 64.48 & 51.12 & 57.84 & 54.04 & 43.44 & \textbf{66.60}\\
    FEC\textsuperscript{\cite{FEC}} (w/ ours)
    & 93.72 & 89.98 & 60.14  & 88.87  & 83.80  & 53.83  & 80.63  & 75.25  & 53.68  & 64.72  & 59.97  & 47.86 & \textbf{71.04}\\
    \rowcolor[rgb]{.9,1,.9}
    \textit{improvement}
    & \textcolor{brickred}{↑1.40} & \textcolor{brickred}{↑3.76} & \textcolor{brickred}{↑1.38}
    & \textcolor{brickred}{↑2.64} & \textcolor{brickred}{↑2.46} & \textcolor{blue}{↓1.2}
    & \textcolor{brickred}{↑12.2} & \textcolor{brickred}{↑10.77} & \textcolor{brickred}{↑2.56}
    & \textcolor{brickred}{↑6.88} & \textcolor{brickred}{↑5.93} & \textcolor{brickred}{↑4.53}
    & \textcolor{brickred}{\textbf{↑4.44}}\\
    \midrule
    ESF\textsuperscript{\cite{ESF}} (w/ OMP\textsuperscript{\cite{OMP}})
    & 89.81 & 85.37 & 57.98 & 91.68 & 88.61 & 54.04 & 89.38 & 85.26 & 56.82 & 75.92 & 71.31 & 50.21 & \textbf{74.70} \\
    ESF\textsuperscript{\cite{ESF}} (w/ ours)
    & 93.02  & 89.78  & 58.13  & 93.84  & 91.27  & 55.22  & 91.66  & 88.04  & 57.01  & 76.68 & 72.74 & 50.82 & \textbf{76.52}  \\
    \rowcolor[rgb]{.9,1,.9}
    \textit{improvement}
    & \textcolor{brickred}{↑3.21} & \textcolor{brickred}{↑4.41} & \textcolor{brickred}{↑0.15}
    & \textcolor{brickred}{↑2.16} & \textcolor{brickred}{↑2.66} & \textcolor{brickred}{↑1.18}
    & \textcolor{brickred}{↑2.28} & \textcolor{brickred}{↑2.78} & \textcolor{brickred}{↑0.19}
    & \textcolor{brickred}{↑0.76} & \textcolor{brickred}{↑1.43} & \textcolor{brickred}{↑0.61}
    & \textcolor{brickred}{\textbf{↑1.82}}\\
    \midrule
    CA-MCNN\textsuperscript{\cite{CA-MCNN}} (w/ OMP\textsuperscript{\cite{OMP}})
    & 94.56 & 92.76 & 49.35 & 93.34 & 91.56 & 51.81 & 90.99 & 86.60  & 52.72 & 80.02 & 73.19 & 44.18 & \textbf{75.09}\\
    CA-MCNN\textsuperscript{\cite{CA-MCNN}} (w/ ours)
    & 96.21  & 94.29  & 52.20 & 94.25 & 92.30  & 54.21  & 90.99 & 86.77  & 55.91 & 79.09 & 72.78  & 44.47 & \textbf{76.12}\\
    \rowcolor[rgb]{.9,1,.9}
    \textit{improvement}
    & \textcolor{brickred}{↑1.65} & \textcolor{brickred}{↑1.53} & \textcolor{brickred}{↑2.85}
    & \textcolor{brickred}{↑0.91} & \textcolor{brickred}{↑0.74} & \textcolor{brickred}{↑2.4}
    & - & \textcolor{brickred}{↑0.17} & \textcolor{brickred}{↑3.19}
    & \textcolor{blue}{↓0.93} & \textcolor{blue}{↓0.41} & \textcolor{brickred}{↑0.29}
    & \textcolor{brickred}{\textbf{↑1.03}}\\
    \midrule
    PAN\textsuperscript{\cite{PAN}} (w/ OMP\textsuperscript{\cite{OMP}})
    & 83.18 & 71.55 & 57.46 & 73.50  & 62.74 & 43.11 & 60.29 & 52.78 & 38.75 & 37.49 & 34.83 & 19.74 & \textbf{52.95}\\
    PAN\textsuperscript{\cite{PAN}} (w/ ours)
    & 91.66 & 84.04  & 56.54 & 85.82 & 77.78 & 42.80 & 75.96  & 71.93  & 36.67 & 54.38 & 50.60 & 29.44  & \textbf{63.13}\\
    \rowcolor[rgb]{.9,1,.9}
    \textit{improvement}
    & \textcolor{brickred}{↑8.48} & \textcolor{brickred}{↑12.49} & \textcolor{blue}{↓0.92}
    & \textcolor{brickred}{↑12.32} & \textcolor{brickred}{↑15.04} & \textcolor{blue}{↓0.31}
    & \textcolor{brickred}{↑15.67} & \textcolor{brickred}{↑19.15} & \textcolor{blue}{↓2.08}
    & \textcolor{brickred}{↑16.89} & \textcolor{brickred}{↑15.77} & \textcolor{brickred}{↑9.70}
    & \textcolor{brickred}{\textbf{↑10.18}}\\
    \midrule
    PIHA\textsuperscript{\cite{PIHA}} (w/ OMP\textsuperscript{\cite{OMP}})
    & 98.22 & 96.22 & 66.10  & 97.59 & 94.36 & 66.47 & 93.41 & 89.31 & 64.88 & 78.56 & 72.49 & 60.23 & \textbf{81.48} \\
    PIHA\textsuperscript{\cite{PIHA}} (w/ ours)
    & 98.52 & 95.91 & 69.41 & 97.42 & 93.91 & 68.55 & 94.24 & 90.65 & 66.26 & 80.37 & 74.43 & 59.41 & \textbf{82.42}\\
    \rowcolor[rgb]{.9,1,.9}
    \textit{improvement}
    & \textcolor{brickred}{↑0.30} & \textcolor{blue}{↓0.31} & \textcolor{brickred}{↑3.31}
    & \textcolor{blue}{↓0.17} & \textcolor{blue}{↓0.45} & \textcolor{brickred}{↑2.08}
    & \textcolor{brickred}{↑0.83} & \textcolor{brickred}{↑1.34} & \textcolor{brickred}{↑1.38}
    & \textcolor{brickred}{↑1.81} & \textcolor{brickred}{↑1.94} & \textcolor{blue}{↓0.82}
    & \textcolor{brickred}{\textbf{↑0.94}}\\
    \bottomrule
    \end{tabular}}
  \label{tab:recon_comparison}
\end{table*}

\begin{figure}[!ht]
\centering
\includegraphics[width=0.9\linewidth]{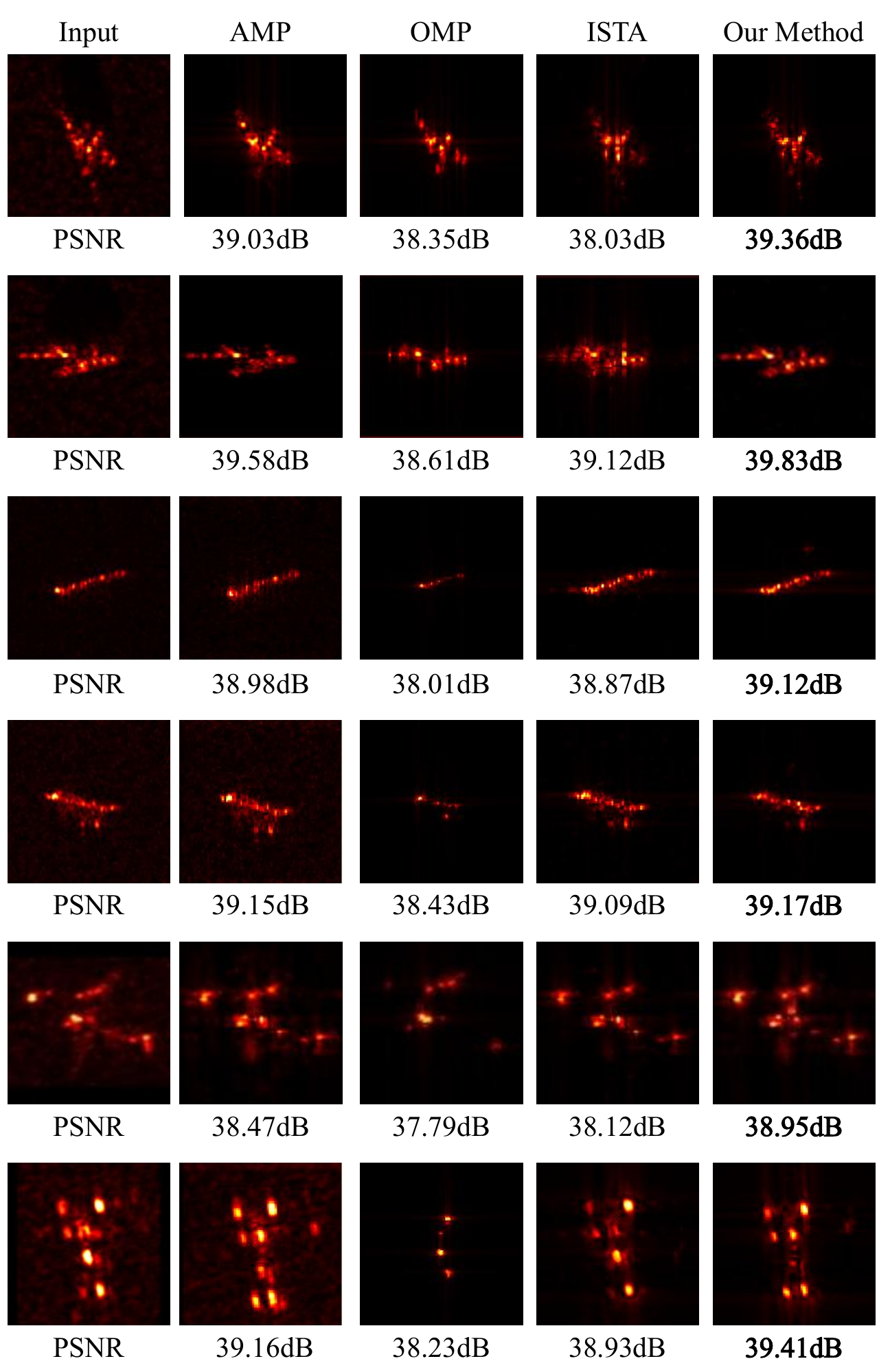}
\caption{Comparison of the reconstructed images and PSNR between ours and traditional methods on MSTAR dataset \cite{mstar}, OpenSARShip dataset \cite{opensarship} and SAR-Aircraft-1.0 dataset \cite{aircraft}. The initial two rows represent MSTAR slices, the subsequent two rows depict OpenSARShip data, and the final two rows correspond to SAR-Aircraft-1.0 targets.} 
\label{fig:recon}
\end{figure}

We evaluate the ESC parameter estimation performance against three conventional approaches: AMP \cite{AMP}, OMP \cite{ori_omp}, and ISTA \cite{beck2009fast}. All methods employ the same dictionary $\Phi$ as defined in Section \ref{sec:ESC}. The hyperparameters are configured as follows: OMP (number of ESC = 40), ISTA ($t$ = 0.01, $\rho$ = 0.005), and AMP (rate of change = 0.01). As shown in Fig. \ref{fig:recon_trad}, our method achieves higher PSNR than those in terms of the reconstructed images. 
The visual comparison in Fig. \ref{fig:recon} further demonstrates the superior capability obtained by ours in reconstructing images with enhanced sparsity, improved clarity, and more complete EM feature representation. The performance advantage stems from ours' ability to automatically learn optimal dictionaries and hyperparameters during optimization, unlike traditional methods that depend on fixed parameters whose suboptimal choices may limit performance across varying scenarios.

We further evaluated several DNN-based SAR-ATR methods incorporating ESC parameters, including FEC \cite{FEC}, ESF \cite{ESF, ESF_2}, CA-MCNN \cite{CA-MCNN}, PAN \cite{PAN}, and PIHA \cite{PIHA}. Table \ref{tab:recon_comparison} presents a comparative analysis of these methods when using ESC parameters extracted by either our method or OMP approach. All experiments were performed on the MSTAR dataset under the OFA evaluation protocols. Replacing the ESC features derived from OMP with those of the proposed approach would introduce significant improvement of recognition performance. The most substantial differences are observed in FEC and PAN, indicating their heightened sensitivity to ESC feature quality. The overall results confirm the superior robustness and generalization capability of ours in handling limited training data scenarios.

\subsubsection{Analysis on Generalization Ability}

\begin{figure}[H]
\centering
\includegraphics[width=0.9\linewidth]{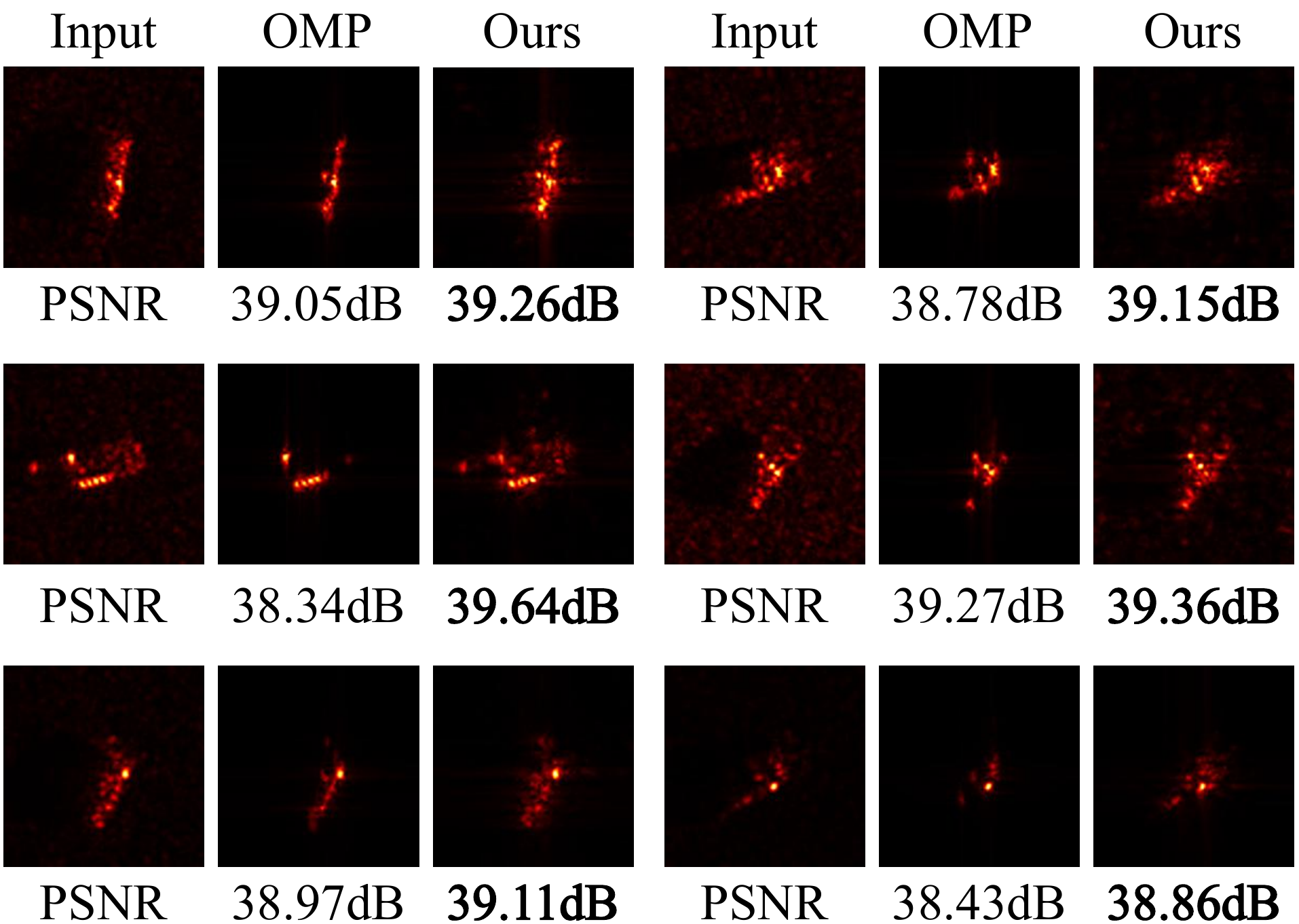}
\caption{Comparison of the reconstructed images and PSNR between ours and OMP \cite{OMP} method on six slices in SAMPLE dataset.} 
\label{fig:sample_recon}
\end{figure}

The proposed approach integrates physics-driven optimization with deep learning, offering inherent interpretability while maintaining strong performance with limited training data. To evaluate its generalization capability, we conduct a synthetic-to-real experiment using the SAMPLE dataset, where the network is trained on only five synthetic images per category and tested on real SAR data. As demonstrated in Figs. \ref{fig:recon_trad} and \ref{fig:sample_recon}, the proposed method achieves substantially better ESC parameter estimation compared to conventional methods while effectively bridging the synthetic-to-real domain gap. 

Furthermore, Table \ref{tab:sample_asc} shows that the estimated ESC features by ours consistently enhance performance across multiple deep learning recognition methods compared to OMP, validating its robust generalization capability across different data domains under extreme data scarcity conditions.

\begin{table}[!htbp]
  \centering
  \small
  \caption{Performances of various SAR target recognition approaches that combine DNNs and ESC features obtained from the proposed method and other traditional optimization methods. The experiments are conducted on SAMPLE dataset (synthetic-to-real scenario).}
  \resizebox{1.0\linewidth}{!}{%
    \begin{tabular}{ccccc}
      \toprule
      \rowcolor[rgb]{.929,.929,.929}
      \textbf{Method} & \textbf{10\%} & \textbf{30\%} & \textbf{50\%} & \textbf{100\%} \\
      \midrule
      FEC\textsuperscript{\cite{FEC}} (w/ OMP\textsuperscript{\cite{OMP}}) & 54.09 & 61.32 & 63.9  & 70.55 \\
      FEC\textsuperscript{\cite{FEC}} (w/ ours)                            & 62.95 & 65.86 & 67.42 & 75.88 \\
      \rowcolor[rgb]{.9,1,.9} \textit{improvement} & \textcolor{brickred}{↑8.86} & \textcolor{brickred}{↑4.54} & \textcolor{brickred}{↑3.52} & \textcolor{brickred}{↑5.33} \\
      \midrule
      ESF \textsuperscript{\cite{ESF}} (w/ OMP\textsuperscript{\cite{OMP}}) & 63.67 & 78.78 & 84.79 & 85.33 \\
      ESF \textsuperscript{\cite{ESF}} (w/ ours)                             & 65.88 & 79.53 & 84.91 & 86.5 \\
      \rowcolor[rgb]{.9,1,.9} \textit{improvement} & \textcolor{brickred}{↑2.21} & \textcolor{brickred}{↑0.75} & \textcolor{brickred}{↑0.12} & \textcolor{brickred}{↑1.17} \\
      \midrule
      CA-MCNN \textsuperscript{\cite{CA-MCNN}} (w/ OMP\textsuperscript{\cite{OMP}}) & 40.97 & 46.77 & 46.55 & 40.5 \\
      CA-MCNN \textsuperscript{\cite{CA-MCNN}} (w/ ours)                          & 41.07 & 45.78 & 49.85 & 46.85 \\
      \rowcolor[rgb]{.9,1,.9} \textit{improvement} & \textcolor{brickred}{↑0.10} & \textcolor{blue}{↓0.99} & \textcolor{brickred}{↑3.30} & \textcolor{brickred}{↑6.35} \\
      \midrule
      PAN \textsuperscript{\cite{PAN}} (w/ OMP\textsuperscript{\cite{OMP}}) & 42.38 & 54.76 & 63.75 & 73.65 \\
      PAN \textsuperscript{\cite{PAN}} (w/ ours)                           & 43.00 & 64.57 & 73.05 & 79.80 \\
      \rowcolor[rgb]{.9,1,.9} \textit{improvement} & \textcolor{brickred}{↑0.62} & \textcolor{brickred}{↑9.81} & \textcolor{brickred}{↑9.30} & \textcolor{brickred}{↑6.15} \\
      \midrule
      PIHA \textsuperscript{\cite{PIHA}} (w/ OMP\textsuperscript{\cite{OMP}}) & 65.01 & 76.9  & 81.14 & 80.52 \\
      PIHA \textsuperscript{\cite{PIHA}} (w/ ours)                           & 67.20 & 78.43 & 81.81 & 82.93 \\
      \rowcolor[rgb]{.9,1,.9} \textit{improvement} & \textcolor{brickred}{↑2.19} & \textcolor{brickred}{↑1.53} & \textcolor{brickred}{↑0.67} & \textcolor{brickred}{↑2.41} \\
      \bottomrule
    \end{tabular}%
  }
  \label{tab:sample_asc}
\end{table}

\subsubsection{Analysis on Efficiency}

Table \ref{tab:inference_time_comparison} presents a comparative analysis of ESC parameter estimation time between the proposed methods and conventional methods, with single-image processing time as the evaluation metric. All traditional methods (OMP, AMP, and ISTA) are implemented in PyTorch with CUDA acceleration. The results reveal that OMP requires a minimum of 100 seconds per image, while AMP and ISTA also exhibit considerable computational demands. In contrast, the proposed method achieves real-time processing at 0.1 seconds per image---representing a three-order-of-magnitude speed improvement. This dramatic computational efficiency not only demonstrates the practical superiority of our approach but also facilitates its direct incorporation into end-to-end neural network frameworks.

\begin{table}[H]
\centering
\caption{Comparison of inference time between traditional methods and our approach across multiple datasets. The best results are highlighted in \textbf{bold}.}
\resizebox{0.9\linewidth}{!}{\begin{tabular}{lcccccc}
\toprule
\multirow{2}{*}{Dataset} & \multicolumn{4}{c}{Inference Time (in seconds) } \\
                         & AMP & OMP & ISTA & Ours \\ \midrule
MSTAR                    & 84.536          & 106.481           & 72.163            & \textbf{0.098} \\ 
SAMPLE                   & 86.298          & 101.328           & 70.662            & \textbf{0.103} \\ 
OpenSARship              & 81.251          & 108.633           & 76.216            & \textbf{0.106} \\ 
CSRSDD                   & 79.237          & 109.392          & 74.545            & \textbf{0.093} \\ 
SAR-Aircraft             & 83.684          & 104.839           & 72.527            & \textbf{0.096} \\ 
\midrule
Average Time & 83.001 & 106.135 & 73.223& \textbf{0.099} \\
\bottomrule
\end{tabular}}
\label{tab:inference_time_comparison}
\end{table}

\subsection{Effectiveness on Image Recognition}
\label{TR}

\subsubsection{Comparison with SOTA Lightweight Models}

We first benchmark our KINN model against state-of-the-art lightweight architectures, including both CNN-based (MobileNetV3-Large \cite{MobileNetV3-Large}, ShuffleNetV2 \cite{ShuffleNetV2}, GhostNetV3 \cite{GhostNetV3}, SqueezeNet \cite{SqueezeNet}, EfficientNet-B0 \cite{EfficientNet}, A-ConvNet \cite{Aconv}) and ViT-based (EfficientViT-M0 \cite{EfficientNet}, FastViT \cite{FastViT}, TinyViT \cite{TinyViT}, MobileViT-XS/XXS \cite{MobileViT}, EdgeNeXt-XS/XXS \cite{EdgeNeXt} and EfficientFormerV2 \cite{EfficientFormer}). For fair comparisons, we implement both CNN and ViT variants of KINN under identical experimental conditions.



Table~\ref{tab:mstar} summarizes MSTAR recognition performance under the challenging OFA protocol.  With only 0.7M (KINN-CNN) and 0.95M (KINN-ViT) parameters, our models achieve state-of-the-art performance while being the most compact architectures among all compared lightweight methods. With only 10\% training data, KINN delivers an average 11\% accuracy improvement across all OFA scenarios. Specifically, it outperforms leading lightweight CNNs by 25.06\% (OFA1) and 22.57\% (OFA2), and surpasses ViT counterparts by 4.83\% (OFA1) and 7.32\% (OFA2). Most remarkably, in OFA3---the most demanding test of robustness against depression angle variations---KINN achieves 21.03\% and 9.84\% gains over CNN- and ViT-based approaches respectively, demonstrating exceptional generalization under operational conditions.



\begin{table*}[htbp]
  \centering
  \caption{Comparison of target recognition performance with state-of-the-art (SOTA) lightweight CNN and ViT models on the MSTAR dataset using the OFA evaluation protocol. \textbf{Bold} and \underline{underlined} entries denote the best and second-best results, respectively.}
    \resizebox{0.95\linewidth}{!}{
    \begin{tabular}{c|c|ccc|ccc|ccc|ccc}
    \toprule
    \rowcolor[rgb]{ .929,  .929,  .929} & & \multicolumn{3}{c|}{\textbf{90\%}} & \multicolumn{3}{c|}{\textbf{50\%}} & \multicolumn{3}{c|}{\textbf{30\%}} & \multicolumn{3}{c}{\textbf{10\%}}\\
    \rowcolor[rgb]{ .929,  .929,  .929}  \multirow{-2}{*}{\textbf{Method}}   & \multirow{-2}{*}{\textbf{Param.}} & \textbf{OFA1}  & \textbf{OFA2}  & \textbf{OFA3}  & \textbf{OFA1}  & \textbf{OFA2}  & \textbf{OFA3} & \textbf{OFA1}  & \textbf{OFA2}  & \textbf{OFA3}  & \textbf{OFA1}  & \textbf{OFA2}  & \textbf{OFA3} \\
    \midrule
    MobileNetV3 \cite{MobileNetV3-Large} &4.23M & 92.23 & 89.63 & 37.22 & 74.71 & 68.22 & 39.53 & 61.20  & 55.7  & 35.42 & 42.15 & 39.59 & 32.76\\
    ShuffleNetV2 \cite{ShuffleNetV2} &5.37M & 87.22 & 84.01 & \underline{54.75} & 72.27 & 69.37 & \underline{47.72} & 63.40 & 59.06 & 43.16 & 60.91 & 56.47 & 30.83\\
    GhostNetV3 \cite{GhostNetV3} & 6.86M& 92.73 & 90.62 & 30.55 & 92.28 & 87.71 & 34.71 & 88.22 & 83.67 & 35.36 & 58.61 & 54.12 & 22.13\\
    SqueezeNet \cite{SqueezeNet} & \underline{0.73M}& 84.84 & 80.81 & 40.09 & 80.54 & 75.53 & 38.21 & 74.61 & 69.43 & 42.75 & 61.73 & 56.3 & 30.45\\
    EfficientNet \cite{EfficientNet} &4.02M & \underline{94.00} & \underline{90.65} & 44.28 & \underline{97.7}  & \underline{94.44} & 47.23 & \underline{90.83} & \underline{86.49} & 48.48 & \underline{63.31} & \underline{58.03} & 37.98\\
    ESPNetV2 \cite{espnetv2} & 2.23M& 87.23 & 82.43 & 50.57 & 76.12 & 69.01 & 42.55 & 83.27 & 78.91 & \underline{52.01} & 60.42 & 57.86 & \underline{44.15}\\
    \rowcolor[rgb]{ .9,  1,  .9}  \textbf{KINN-CNN} (ours)  & \textbf{0.70M}& \textbf{97.94} & \textbf{97.69} & \textbf{71.11} & \textbf{97.81} & \textbf{95.97} & \textbf{71.01} & \textbf{96.45} & \textbf{94.82} & \textbf{68.19} & \textbf{88.37} & \textbf{80.60} & \textbf{65.18}\\
    
    \bottomrule
    \toprule

    EfficientViT-M0 \cite{EfficientViT} & 2.16M& \textbf{97.95} & 94.80  & \underline{65.08} & \textbf{97.24} & \textbf{93.59} & 61.37 & \underline{93.07} & 86.11 & \underline{64.33} & 63.61 & 57.53 & 42.17\\
    FastViT \cite{FastViT} & 3.05M & 95.97 & 93.17 & 58.81 & 93.62 & 89.50  & 56.34 & 84.77 & 82.59 & 49.77 & 62.00 & 60.44 & 39.53 \\
    TinyViT-5M \cite{TinyViT} & 5.39M & 97.37 & 94.86 & 63.18 & 95.17 & 89.90  & \underline{63.05} & 89.49 & 83.53 & 52.90  & 64.59 & 59.62 & 32.38 \\
    MobileVIT-XXS \cite{MobileViT} & \underline{0.95M}& 97.36 & 94.28 & 62.12 & 95.65 & 91.71 & 44.40 & 87.38 & 83.38 & 55.75 & 52.03 & 46.76 & 28.96 \\
    EdgeNeXt-XXS \cite{EdgeNeXt} & 1.16M& 95.92 & 91.56 & 55.08 & 93.72 & 91.73 & 62.50  & 90.91 & 86.31 & 56.10  & 64.54 & 59.03 & \underline{53.94} \\
    EfficientFormerV2 \cite{EfficientFormer} &3.42M & \underline{97.79} & \underline{95.51} & 63.30  & 95.81 & 92.51 & 61.97 & 92.18 & \underline{86.52} & 55.28 & \underline{70.68} & \underline{63.78} & 50.51  \\
    \rowcolor[rgb]{ .9,  1,  .9} \textbf{KINN-ViT} (ours) &\textbf{0.95M} & 97.72 & \textbf{96.22} & \textbf{70.36} & \underline{96.71} & \underline{93.35} & \textbf{71.56} & \textbf{94.56} & \textbf{89.23} & \textbf{71.72} & \textbf{75.51} & \textbf{71.10} & \textbf{63.78}\\
    \bottomrule
    \end{tabular}}
  \label{tab:mstar}
\end{table*}

Table~\ref{tab:osr_csrsdd} compares KINN against SOTA lightweight CNN/ViT methods on OpenSARShip, CSRSDD, and SAR-Aircraft-1.0 datasets. Our models achieve top performance in 21 of 22 test scenarios, with KINN-CNN showing 9.57\% and KINN-ViT 7.77\% accuracy gains over respective runner-ups on OpenSARShip's small-scale images. This advantage stems from KINN's unique physics-informed architecture that effectively extracts target features from limited pixel where conventional methods fail. Under 10\% training data conditions, KINN maintains strong performance with average accuracy improvements of 7.73\% (CNN variant) and 3.86\% (ViT variant) on CSRSDD and SAR-Aircraft-1.0, demonstrating consistent robustness across resolution variations and data scarcity conditions.


\begin{table*}[htbp]\small
  \centering
  \setlength{\tabcolsep}{8pt} 
  \caption{Comparison of target recognition performance with state-of-the-art (SOTA) lightweight CNN and ViT models on OpenSARShip, CSRSDD, and SAR-Aircraft-1.0 dataset. \textbf{Bold} and \underline{underlined} entries denote the best and second-best results, respectively.}
    \resizebox{0.95\linewidth}{!}{\begin{tabular}{c|ccc|cccc|cccc}
    \toprule
     \rowcolor[rgb]{ .929,  .929,  .929} & \multicolumn{3}{c|}{\textbf{OpenSARShip}} & \multicolumn{4}{c|}{\textbf{CSRSDD}} & \multicolumn{4}{c}{\textbf{SAR-Aircraft-1.0}} \\
    \rowcolor[rgb]{ .929,  .929,  .929} \multirow{-2}{*}{\textbf{Method}} & \textbf{Small} & \textbf{Mid}   & \textbf{Large} & \textbf{10\%}  & \textbf{30\%}  & \textbf{50\%}  & \textbf{100\%} & \textbf{10\%}  & \textbf{30\%}  & \textbf{50\%}  & \textbf{100\%} \\
    \midrule
    MobileNetV3 \cite{MobileNetV3-Large} & 50.77 & 60.74 & 77.48 & 62.17 & 69.24 & 71.37 & 78.52 & 68.25 & 92.92 & \underline{97.69} & 97.57 \\
    ShuffleNetV2 \cite{ShuffleNetV2} & 50.28 & \underline{62.05} & \underline{82.46} & \underline{64.85} & 68.96 & 72.43 & \underline{79.51} & 69.36 & 92.24 & 97.28 & 98.86  \\
    GhostNetV3 \cite{GhostNetV3} & 51.66 & 58.6  & 82.07 & 64.21 & 69.52 & 73.32 & 78.69 & 68.55 & 91.94 & 95.34 & 97.98 \\
    SqueezeNet \cite{SqueezeNet} & \underline{53.01} & 60.64 & 78.95 & 63.47 & \underline{70.83} & 73.04 & 79.29 & \underline{71.47} & \underline{93.66} & 96.99 & \underline{99.04} \\
    EfficientNet \cite{EfficientNet} & 52.50  & 60.39 & 81.24 & 63.69 & 70.14 & \underline{73.91} & 78.63 & 67.62 & 91.03 & 95.32 & 97.94  \\
    ESPNetV2 \cite{espnetv2} & 49.73 & 60.05 & 80.82 & 51.22 & 66.73 & 70.27 & 75.83 & 69.44&92.36&97.46&98.91\\
    \rowcolor[rgb]{ .9,  1.0,  .9} \textbf{KINN-CNN} (ours)  & \textbf{62.58} & \textbf{69.56} & \textbf{82.76} & \textbf{70.65} & \textbf{74.38} & \textbf{78.17} & \textbf{81.63} & \textbf{80.86} & \textbf{95.76} & \textbf{98.71} & \textbf{99.79} \\
    
    \bottomrule
    \toprule
   
    EfficientViT-M0 \cite{EfficientViT} & 50.25 & 60.22 & 82.00  & 64.38 & 62.05  & \underline{72.62} & 72.51 & 68.52 & 92.40 & \underline{97.13} & 99.01  \\
    FastViT \cite{FastViT} & \underline{52.77} & 60.59 & 80.09 & 62.24 & 67.23 & 70.25 & 76.00 & 65.83 & 90.77 & 95.93 & 98.66 \\
    TinyViT \cite{TinyViT} & 50.32 & \underline{60.78} & 82.07 & 60.42 & 67.61 & 72.29 & \textbf{77.32} & \underline{69.48} & \underline{92.57} & 96.97 & \underline{99.15} \\
    MobileVIT-XXS \cite{MobileViT} & 51.41 & 59.71 & 80.90  & 61.72 & 59.81 & 71.01 & 74.13 & 66.58 & 92.11 & 96.86 & 98.84\\
    EdgeNeXt-XXS \cite{EdgeNeXt} & 51.13 & 60.71 & 81.31 & 53.68  & 56.06 & 66.39 & 71.93 & 67.60 & 92.31 & 96.79 & 98.77\\
    EfficientFormerV2 \cite{EfficientFormer} & 49.04 & 60.20  & \underline{82.46} & \underline{67.48} & \underline{69.31} & 69.98 & 72.94  & 68.02 & 92.05 & 97.08 & 98.78\\
    \rowcolor[rgb]{ .9,  1.0,  .9} \textbf{KINN-ViT} (ours) & \textbf{60.64} & \textbf{61.33} & \textbf{82.51} & \textbf{69.01} & \textbf{71.81} & \textbf{74.63} & \underline{76.98}  & \textbf{75.67} & \textbf{92.68} & \textbf{98.86} & \textbf{99.51}\\
    \bottomrule
    \end{tabular}}
  \label{tab:osr_csrsdd}
\end{table*}

\subsubsection{Comparison with SOTA SAR-ATR Methods}


Table \ref{tab:4.3.2} compares KINN against state-of-the-art SAR-ATR methods on MSTAR datasets under the OFA protocol. The comparison includes both physics-aware approaches (FEC, ESF, CA-MCNN, PAN, PIHA) that leverage SAR-specific electromagnetic features for enhanced generalization, and data-driven networks (A-ConvNet for amplitude data, MSNet for complex data). KINN demonstrates superior accuracy across all test conditions while maintaining competitive inference speed and number of learnable parameters, establishing its advantages in both generalization and computational efficiency.

\begin{table*}[htbp]
  \centering
  \setlength{\tabcolsep}{4pt}
  \caption{Comparison of the learnable parameters (Param.), frames per second (FPS) and target recognition performance with state-of-the-art (SOTA) SAR-ATR methods on MSTAR dataset. \textbf{Bold} and \underline{underlined} entries denote the best and second-best results, respectively.}
  \resizebox{1.0\linewidth}{!}{
    \begin{tabular}{c@{\hskip 1pt}|cc|ccc|ccc|ccc|ccc|c}
      \toprule
      \rowcolor[rgb]{.929,.929,.929}
       &
       &
       &
      \multicolumn{3}{c|}{\textbf{90\%}} &
      \multicolumn{3}{c|}{\textbf{50\%}} &
      \multicolumn{3}{c|}{\textbf{30\%}} &
      \multicolumn{3}{c|}{\textbf{10\%}} &
       \\
      \rowcolor[rgb]{.929,.929,.929}
      \multirow{-2}{*}{\textbf{Method}} & \multirow{-2}{*}{\textbf{Param.}}&\multirow{-2}{*}{\textbf{FPS}} &
      \textbf{OFA1} & \textbf{OFA2} & \textbf{OFA3} &
      \textbf{OFA1} & \textbf{OFA2} & \textbf{OFA3} &
      \textbf{OFA1} & \textbf{OFA2} & \textbf{OFA3} &
      \textbf{OFA1} & \textbf{OFA2} & \textbf{OFA3} & \multirow{-2}{*}{\textbf{Average}}\\
      \midrule
      FEC \cite{FEC} & 16.82M & 31.8
      & 93.72 & 89.98 & 60.14  & 88.87  & 83.80  & 53.83  & 80.63  & 75.25  & 53.68  & 64.72  & 59.97  & 47.86 & 71.04\\
      ESF \cite{ESF} & \underline{0.22M} & \underline{100}
      & 93.02  & 89.78  & 58.13  & 93.84  & 91.27  & 55.22  & 91.66  & 88.04  & 57.01  & 76.68 & 72.74 & 50.82 & 76.52  \\
      CA-MCNN \cite{CA-MCNN} & 4.96M & 77.2
      & 96.21  & 94.29  & 52.20 & 94.25 & 92.30  & 54.21  & 90.99 & 86.77  & 55.91 & 79.09 & 72.78  & 44.47 & 76.12\\
      PAN \cite{PAN} & 1.82M & 97
      & 91.66 & 84.04  & 56.54 & 85.82 & 77.78 & 42.80 & 75.96  & 71.93  & 36.67 & 54.38 & 50.60 & 29.44  & 63.13\\
      PIHA \cite{PIHA} & 18.56M & 54.8
      & 94.82 & \underline{95.91} & \underline{69.41} & 97.42 & 93.91 & \underline{68.55} & \underline{94.24} & \underline{90.65} & \underline{66.26} & \underline{80.37} & \underline{74.43} & 59.41 & \underline{82.42}\\
      A-ConvNet \cite{Aconv} & \textbf{0.08M} & 74.60
      & 86.95 & 84.51 & 57.16 & 92.48 & 88.88 & 62.80 & 87.65 & 83.04 & 58.63 & 72.02 & 64.76 & 52.71 & 74.30\\
      MSNet \cite{MS-CVNets} & 16.75M & 50.25
      & \textbf{97.97} & 95.33 & 64.20 & \underline{97.44} & \underline{94.41} & 65.84 & 92.94 & 88.33 & 63.90 & 78.43 & 72.81 & \underline{59.82} & 80.95 \\
      \rowcolor[rgb]{.9,1,.9}
      \textbf{KINN-CNN} (ours) & 0.70M & \textbf{126.24}
      & \underline{97.94} & \textbf{97.69} & \textbf{71.11} & \textbf{97.81} & \textbf{95.97} & \textbf{71.01} & \textbf{96.45} & \textbf{94.82} & \textbf{68.19} & \textbf{88.37} & \textbf{80.60} & \textbf{65.18} & \textbf{85.42}\\
      \bottomrule
    \end{tabular}}
  \label{tab:4.3.2}
\end{table*}

\subsubsection{Analysis on Generalization Ability}


We rigorously evaluate KINN's generalization capability across four challenging scenarios: \textbf{1) Target type variation:} MSTAR OFA2 protocol testing cross-type domain adaptation, \textbf{2) Imaging angle variation:} MSTAR OFA3 protocol evaluating depression angle robustness, \textbf{3) Synthetic-to-real transfer:} SAMPLE dataset assessing performance on real SAR data when trained on synthetic, and \textbf{4) Limited data scenarios:} analyzing performance degradation under progressively reduced training samples.


Table \ref{tab:4.3.3} compares KINN-CNN's performance against state-of-the-art lightweight CNNs and specialized SAR-ATR methods across MSTAR and SAMPLE datasets. KINN-CNN consistently outperforms all competitors, achieving a 6.65\% accuracy improvement with only 10\% training data on MSTAR, and maintaining 3.54\% (OFA2) and 2.96\% (OFA3) advantages in cross-domain scenarios. On SAMPLE, it shows superior generalization with 1.91\% improvement using 10\% training data, and over 5\% gains at higher training proportions (30\%-100\%). These results demonstrate KINN's robust feature learning capability under both data-limited and cross-domain conditions.

\begin{table*}[htbp]\small
  \centering
  \setlength{\tabcolsep}{10pt} 
  \caption{Comparison of target recognition performance using SOTA lightweight CNN and SAR-ATR methods on SAMPLE and MSTAR datasets within various test scenarios. These scenarios included target type generalization (OFA2 protocol of MSTAR), imaging angle generalization (OFA3 protocol of MSTAR), synthetic-to-real generalization (results on SAMPLE dataset), and limited training data generalization. \textbf{Bold} and \underline{underlined} entries denote the best and second-best results, respectively.}
    \resizebox{0.95\linewidth}{!}{
    \begin{tabular}{c|ccc|cc|cccc}
    \toprule
     \rowcolor[rgb]{ .929,  .929,  .929} & \multicolumn{5}{c|}{\textbf{MSTAR (Average Accuracy)}} & \multicolumn{4}{c}{\textbf{SAMPLE}} \\
    \rowcolor[rgb]{ .929,  .929,  .929} \multirow{-2}{*}{\textbf{Method}} & \textbf{10\%}  & \textbf{30\%}  & \textbf{50\%}  & \textbf{OFA2} & \textbf{OFA3}  & \textbf{10\%}  & \textbf{30\%}  & \textbf{50\%}& \textbf{100\%} \\
    \midrule
    MobileNetV3 \cite{MobileNetV3-Large} & 38.17 & 50.77 & 60.82 & 63.29 & 36.23 & 33.32 & 54.02 & 51.14&69.70 \\
    ShuffleNetV2 \cite{ShuffleNetV2} & 49.40 & 55.21 & 63.12 & 67.23 & 44.12 & 65.06 & 79.60 & 80.46 & 82.38\\
    GhostNetV3 \cite{GhostNetV3} & 44.95 & 69.08 &71.57 & 79.03 & 30.69 & 60.82 & 71.05 &74.81 & 77.59\\
    SqueezeNet \cite{SqueezeNet} & 49.49 & 62.26 & 64.76 & 70.52 & 37.86 & 38.49 & 49.16 & 48.17 & 54.77 \\
    EfficientNet \cite{EfficientNet} & 53.11 & 75.27 & 79.79 & 82.40 & 44.49 & 63.00 & 80.37 & 81.27 & 84.67 \\
    ESPNetV2 \cite{espnetv2} & 54.14 & 71.40 & 62.56 & 72.05 & 47.32 & 38.24 & 52.98 &46.50 & 63.25\\
    \midrule
    FEC \cite{FEC}  & 57.52 & 69.85 & 75.50 & 77.25 & 53.88 & 62.95 & 65.86 & 67.42 & 75.88\\
    ESF \cite{ESF}  & 66.75 & 78.90 & 80.11 & 85.46 & 55.30 & 65.88 & 79.53 & \underline{84.91} & \underline{86.50} \\
    CA-MCNN \cite{CA-MCNN}  & 65.45 & 77.89 & 80.25 & 86.54 & 51.70 & 41.07 & 45.78 & 49.85 & 46.85\\
    PAN \cite{PAN}  & 44.81 & 61.52 & 68.80 & 71.09 & 41.36 & 43.00 & 64.57 &73.05 & 79.80\\
    PIHA \cite{PIHA}  & \underline{71.40} & \underline{83.72} & \underline{86.63} & \underline{88.73} & \underline{65.91} & \underline{67.20} & \underline{78.43} & 81.81 & 82.93 \\
    \midrule
    \rowcolor[rgb]{ .9,  1.0,  .9} \textbf{KINN-CNN} (ours) & \textbf{78.05} & \textbf{
86.49} & \textbf{88.26} &\textbf{92.27} & \textbf{68.87} & \textbf{69.11} & \textbf{83.43} & \textbf{86.97}& \textbf{87.61}\\
    \bottomrule
    \end{tabular}}
  \label{tab:4.3.3} 
\end{table*}


\begin{figure*}[!ht]
\centering
\includegraphics[width=0.9\linewidth]{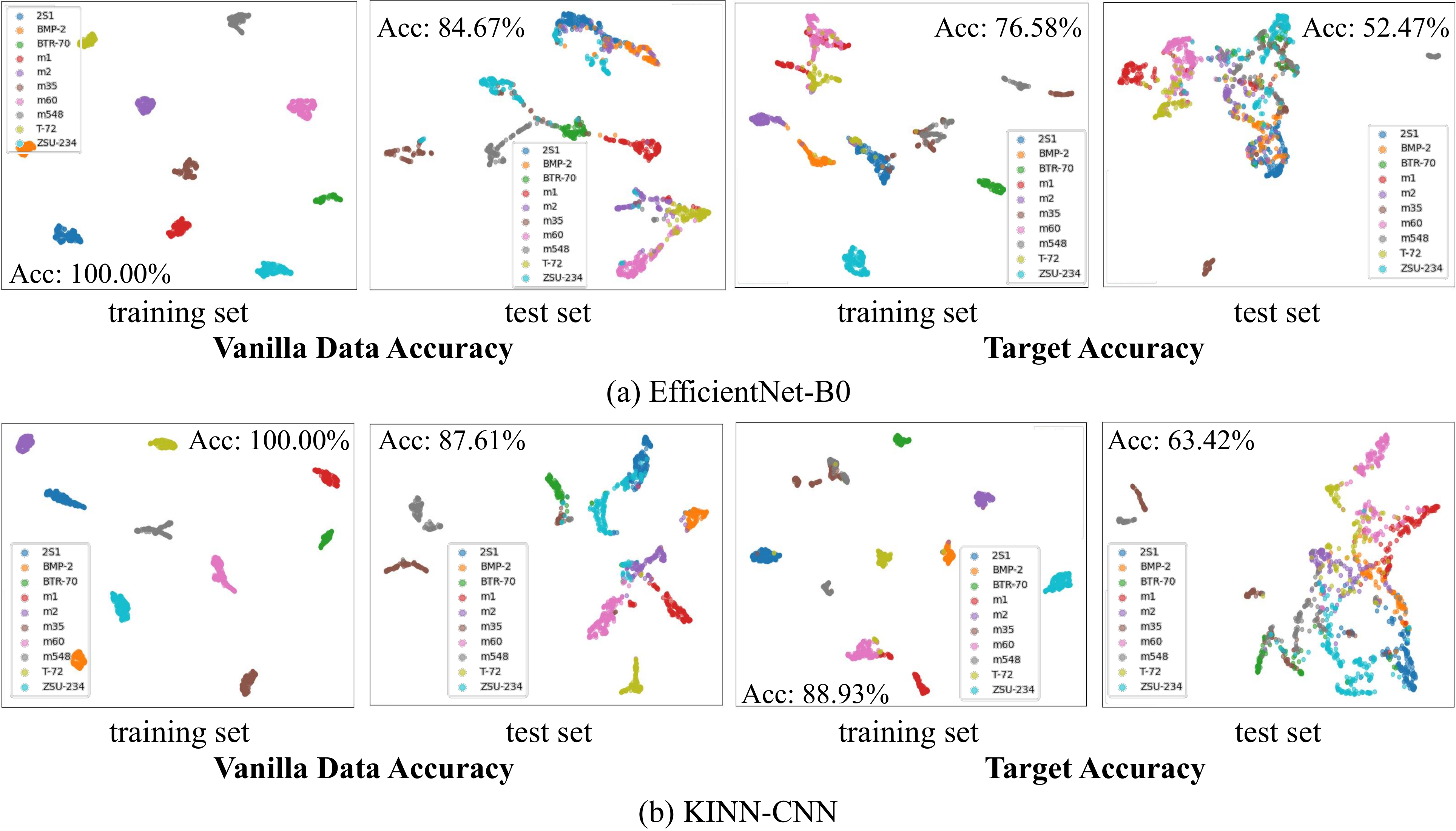}
\caption{UMAP \cite{umap} visualization of (a) EfficientNet-B0 and (b) KINN-CNN on the vanilla data (left column) and target regions only (right column).} 
\label{fig:umap_mask}
\end{figure*}

To evaluate target-specific feature compression, we compare UMAP embeddings \cite{umap} from EfficientNet-B0 \cite{EfficientNet} and KINN-CNN on both vanilla and target-only data across training and test datasets (Fig. \ref{fig:umap_mask}). On vanilla data, both models acquire distinguishable characteristics from the training set; however, KINN-CNN exhibits more favorable retention of cluster structure and separation in the test set, suggesting enhanced feature generalization overall. The essential differentiation arises in the target-only scenario. EfficientNet-B0 \cite{EfficientNet}, although it derives a certain structure from the training data based exclusively on target pixels, reveals significant degradation in feature space organization on the test set, characterized by diffuse and ineffectively separated clusters. This indicates that its compression of target information is not universally applicable. In contrast, KINN-CNN creates highly compact and well-separated clusters utilizing solely target information from the training set, while importantly preserving this superior discriminative structure in the test set. It demonstrates KINN-CNN's enhanced ability to efficiently distill, compress, and generalize the inherent, class-discriminative information exclusively from the target object.

\subsection{Ablation Studies}

\subsubsection{Compression in complex domain}

Table~\ref{tab:abl_AM_combined} shows an ablation study of the key modules in the compression in complex domain on MSTAR and OpenSARShip datasets. As the depression angle annotations are unavailable in the OpenSARShip dataset, the performance of the angle embedding module cannot be evaluated; the results are therefore omitted. The baseline yields the lowest performance (MSTAR: 39.08 PSNR, 87.72\% accuracy; OpenSARShip: 38.98 PSNR, 82.23\% accuracy). Adopting the diagonal shear module slightly achieves improvements, suggesting implicit scattering-center refinement. Incorporating Gaussian Random Matrix (GRM) embedding provides modest gains (MSTAR: 39.47 PSNR, 88.09\% accuracy), though limited by its random angle representation. The proposed angle embedding module achieves the best performance by explicitly encoding depression angles, significantly enhancing parameter estimation and classification accuracy. These results demonstrate that explicit angle information provides superior feature quality compared to implicit or random approaches.

\begin{table}[htbp]
  \centering
  \caption{Ablation study in the physics-guided compression stage. Impact on PSNR and Accuracy (Acc.) (DS: Diagonal Shear, AE: Angle-embedding, GRM: Gaussian Random Matrix).}
  \resizebox{0.75\linewidth}{!}{
    \begin{tabular}{cccccc}
    \toprule
    & & \multicolumn{2}{c}{MSTAR} & \multicolumn{2}{c}{OpenSARShip} \\
    \cmidrule{3-6}
    \multirow{-2}{*}{DS} & \multirow{-2}{*}{AE} & PSNR & Acc. & PSNR & Acc. \\
    \midrule
    \XSolidBrush     & \XSolidBrush     & 39.08 & 87.72 & 38.98 & 82.23 \\
    \CheckmarkBold    & \XSolidBrush     & 39.35 & 87.98 & 39.37 & 82.76 \\
    \CheckmarkBold    & GRM & 39.47 & 88.09 & - & - \\
    \CheckmarkBold    & Ours  & 39.82 & 88.37 & - & - \\
    \bottomrule
    \end{tabular}
  }
  \label{tab:abl_AM_combined}
\end{table}


We evaluate the trade-off between performance (PSNR and accuracy) and efficiency (inference time) across different numbers of unfolding stages (\textit{N}), with the results summarized in Table \ref{tab:abl_stage}. While PSNR gains plateau beyond \textit{N}=3, inference time continues rising. Accuracy generally improves with \textit{N} but shows slight degradation at \textit{N}=5. Despite faster inference at \textit{N}=2, we selected \textit{N}=3 for subsequent experiments as it optimally balances all three metrics.

\begin{table}[htbp]
  \centering
  \caption{The PSNR, inference time and the recognition accuracy with various $N$.}
    \begin{tabular}{cccccc}
    \toprule
    \rowcolor[rgb]{ .929,  .929,  .929} $N$ & 2     & 3     & 4     & 5     & 6 \\
    \midrule
    PSNR(dB) & 39.31 & 39.82 & 39.85 & 39.79 & 39.96 \\
    Inference Time   & 0.0925 & 0.0979 & 0.1303 & 0.1394 & 0.1565 \\
    Accuracy & 88.32 & 88.37 & 88.39 & 88.35  & 88.41\\
    \bottomrule
    \end{tabular}%
  \label{tab:abl_stage}%
\end{table}%

\begin{figure}[!ht]
\centering
\includegraphics[width=\linewidth]{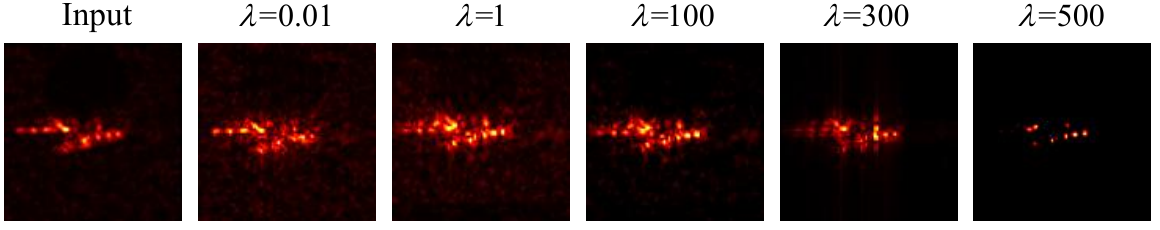}
\caption{The reconstruction image within various $\lambda$.} 
\label{fig:abl_recon}
\end{figure}

As shown in Fig. \ref{fig:abl_recon}, we evaluate the reconstruction results under different settings of $\lambda$. Image quality improves with increasing $\lambda$ up to 300, where background noise is effectively suppressed. Beyond this point ($\lambda$=500), excessive optimization degrades results. We therefore fix $\lambda=300$ for optimal performance in subsequent experiments.

\subsubsection{Aggregation Stage}  

We analyzed the influence of each sparse vector from the reconstruction module and the input image on recognition accuracy, as shown in Table \ref{tab:abl_fuse}. 
The results demonstrate that the model has the greatest impact on the final recognition accuracy when the original images and $z^{(2)}$ are considered, while the contributions of $z^{(1)}$ and $z^{(3)}$ gradually approach zero. This suggests that, in target recognition, the highest performance does not require optimal ESC extraction. Rather, intermediate values play a crucial role in enhancing recognition accuracy. This insight could inspire future research in target recognition using the ESC model.

\begin{table}[htbp]
  \centering
  \setlength{\tabcolsep}{10pt}
  \caption{Ablation study on the contribution of input components in the Aggregation Stage.}
    \resizebox{0.7\linewidth}{!}{\begin{tabular}{llllc}
    \toprule
    \rowcolor[rgb]{ .929,  .929,  .929} $\mathbf{s}$     & $z^{(1)}$    & $z^{(2)}$    & $z^{(3)}$    & Accuracy \\
    \midrule
    \XSolidBrush     &       &       &       &  84.31\\
          & \XSolidBrush     &       &       &  87.97\\
          &       & \XSolidBrush     &       &  85.34\\
          &       &       & \XSolidBrush     &  88.02\\
    \midrule
    \CheckmarkBold     & \CheckmarkBold     & \CheckmarkBold     & \CheckmarkBold     & 88.37 \\
    \bottomrule
    \end{tabular}}
  \label{tab:abl_fuse}
\end{table}

\subsubsection{Ablation of the Overall Framework} 

Table \ref{tab:abl_overall} presents an ablation study of our framework against the MSNet baseline \cite{MS-CVNets}. Incorporating the deep unfolding network (DUN) significantly improves accuracy by leveraging sparse ESC features. This is further enhanced by the angle embedding and diagonal shear modules, which use physical priors to guide information extraction. While replacing standard convolutions with DWConv reveals a trade-off between efficiency and representational capacity, the final addition of self-distillation (SD) markedly improves generalization, which is attributed to SD's ability to enforce semantic compression, aligning intermediate features with the final task.

\begin{table}[htbp]
  \centering
  \setlength{\tabcolsep}{6pt}
  \caption{Ablation study of the overall framework (DS: Diagonal Shear Module, AE: Angle embedding Module, DUN: ISTA-based Unfolding Network, DWConv: Depthwise Convolution, SD: Self-Distillation).}
    \begin{tabular}{ccccc}
    \toprule
    \rowcolor[rgb]{ .929,  .929,  .929}  DS \& AE & DUN  & DWConv & SD & Accuracy \\
    \midrule
    \XSolidBrush & \XSolidBrush & \XSolidBrush & \XSolidBrush &  78.43\\
    \XSolidBrush & \CheckmarkBold  & \XSolidBrush & \XSolidBrush &  84.41\\
    \CheckmarkBold  & \CheckmarkBold  & \XSolidBrush & \XSolidBrush &  85.06\\
    \CheckmarkBold  & \CheckmarkBold  & \CheckmarkBold  & \XSolidBrush &  84.87\\
    \CheckmarkBold  & \CheckmarkBold  & \CheckmarkBold  & \CheckmarkBold  &  88.37\\
    \bottomrule
    \end{tabular}
  \label{tab:abl_overall}%
\end{table}%

\subsection{Discussions and Explanations}

To support our claim that KINN more efficiently compresses SAR data into low-dimensional representations by incorporating domain knowledge, we conduct a series of in-depth experiments. 

\subsubsection{Discussion Based on Information Bottleneck Theory}

The information bottleneck theory posits that training DNNs involves compressing input data into representations that retain minimal mutual information with the input while preserving maximal mutual information relevant to the task label. Optimal representations are those that efficiently capture task-relevant information while discarding redundant or irrelevant details. Patel et al. \cite{patel2024learning} introduced the concept of local rank for individual layers in a DNN to quantify the dimensionality of feature manifolds. A lower local rank signifies greater information compression within a layer. According to \cite{patel2024learning}, the local rank of a layer $l$ has an upper bound that is directly proportional to the norm of the layer weights. As a result, a reduced upper bound of the local rank in a layer implies improved information compression, thereby enhancing the effectiveness of the learned representations.

\begin{figure}[!ht]
\centering
\includegraphics[width=0.95\linewidth]{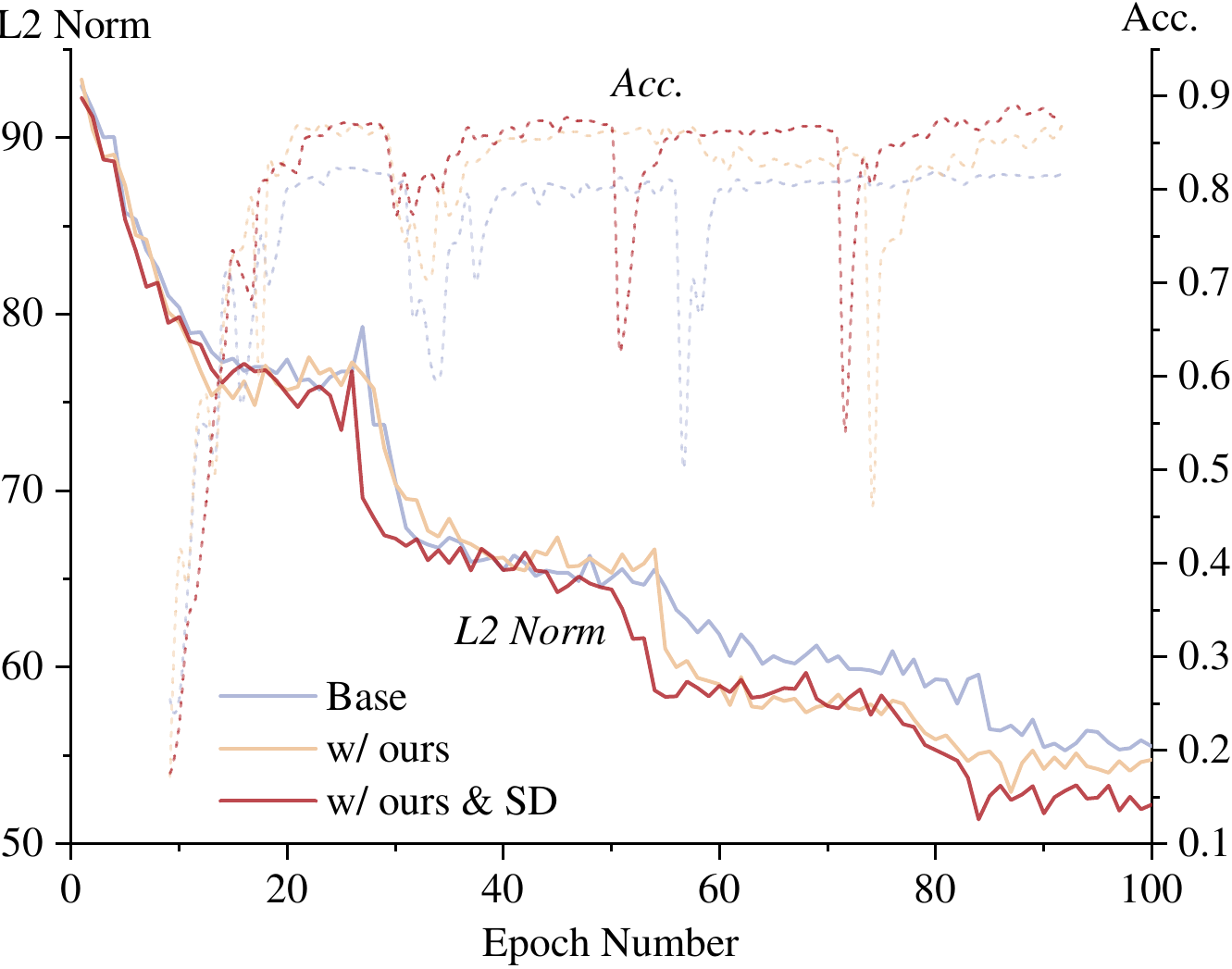}
\caption{The L2 norm of the features from the third CNN block and the accuracy employing various training procedures.} 
\label{fig:L2norm}
\end{figure}


Fig.~\ref{fig:L2norm} analyzes the corresponding accuracy and the feature compression of the third CNN block from the recognition backbone using L2 norm across three training configurations: \textit{Base} (\textit{w/o ours} \& \textit{SD}), \textit{w/ ours}, and our full method \textit{w/ ours} \& \textit{SD}. While all start similarly, the proposed method leads to a rapid reduction in the L2 norm, achieving the most substantial compression. The application of SD offers further refinement, although its contribution shows diminishing returns when compared to the dominant impact of our proposed method. The presented results substantiate the capability of the proposed method to derive compact and discriminative representations.

\subsubsection{Explanation of Knowledge Point}

\begin{figure}[!ht]
\centering
\includegraphics[width=0.8\linewidth]{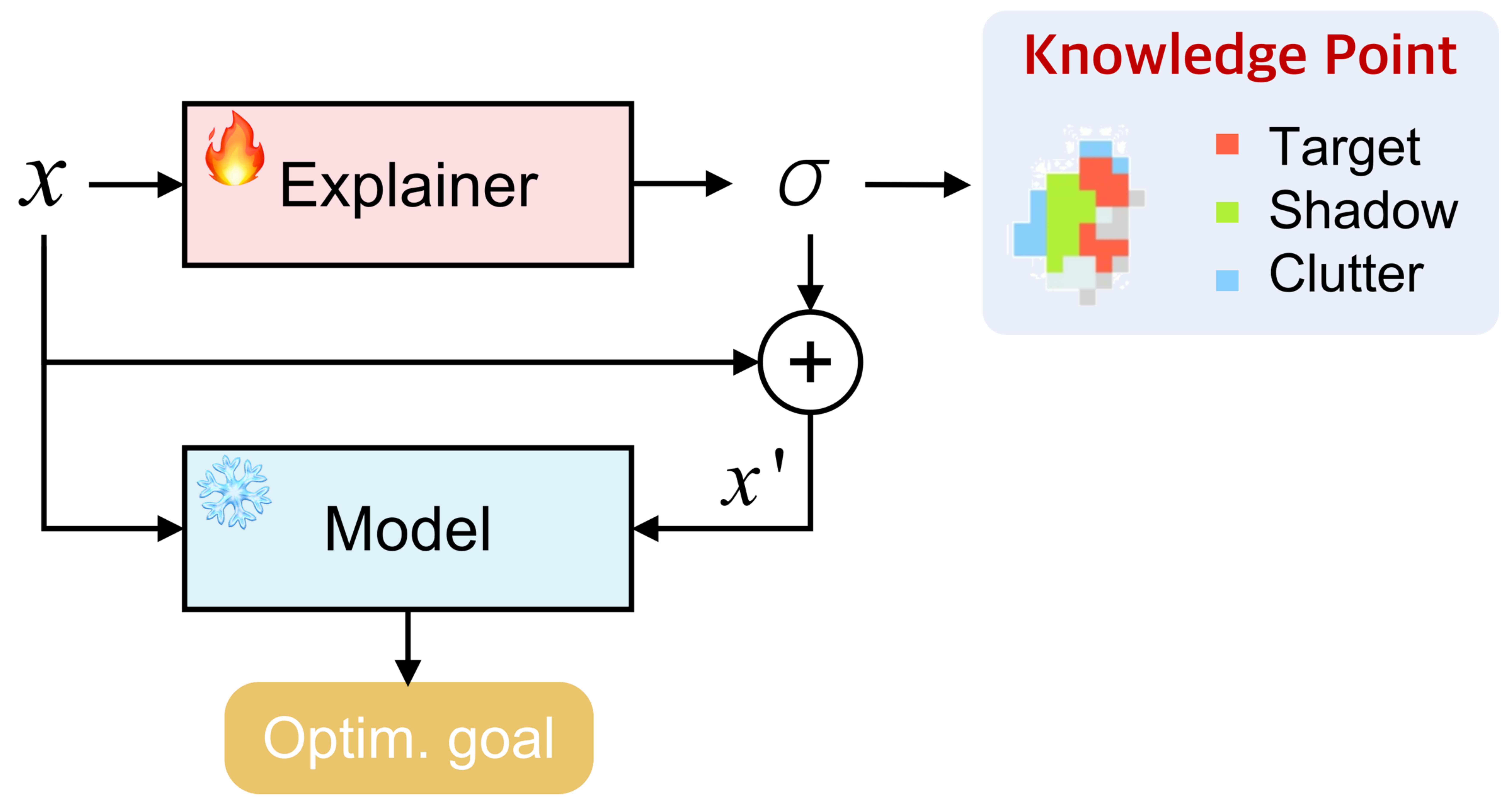}
\caption{A simplified schematic diagram of Knowledge Point (KP) explainer \cite{KP}.} 
\label{fig:KPscheme}
\end{figure}

We analyze feature encoding using the Knowledge Point (KP) method \cite{zhang2022quantifying, KP}, illustrated in Fig. \ref{fig:KPscheme}. The KP explainer quantifies the most influential input information by optimizing perturbations to minimize feature embedding differences between original and perturbed images. This reveals what the model retains during training. By categorizing KPs into target-, shadow-, and clutter-related groups via SAR segmentation, we evaluate model performance: superior models exhibit more target-related KPs and fewer clutter-related KPs, demonstrating effective encoding of relevant features while suppressing noise.

\begin{figure*}[!ht]
\centering
\includegraphics[width=0.8\linewidth]{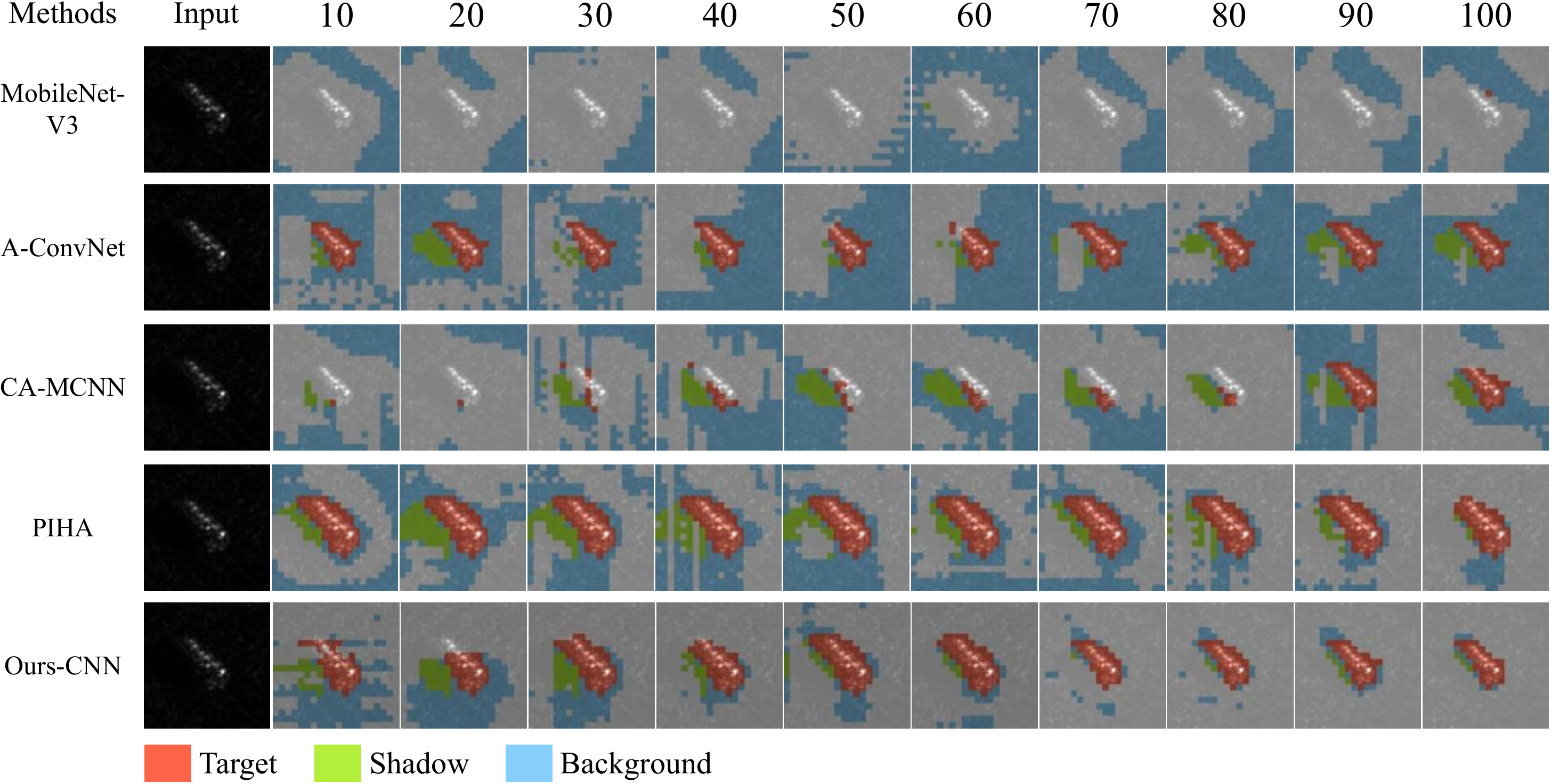}
\caption{A visual comparison of the knowledge points for two lightweight models, two SOTA methods for SAR-ATR, and our model at every 10th epoch during 100 training epochs, with the red, green, and blue areas representing the knowledge points in target, shadow, and background regions, respectively} 
\label{fig:KP_result}
\end{figure*}

We evaluate model effectiveness in encoding critical features by analyzing KP explanations (Fig. \ref{fig:KP_result}) at 10-epoch intervals, using final-layer embeddings from the models trained on 10\% of MSTAR data. Target, shadow, and background regions are marked by red, green, and blue squares, respectively. MobileNetV3-Large \cite{MobileNetV3-Large} shows poor target encoding early on, while CA-MCNN \cite{CA-MCNN} captures target features only in later stages but retains excessive background/shadow interference. A-ConvNet \cite{Aconv} and PIHA \cite{PIHA} quickly learn target representations but overemphasize non-target regions, compromising stability. In contrast, KINN demonstrates superior control: within the first 30 epochs, it efficiently encodes target-related KPs via the sparse representations, then systematically prunes irrelevant features through self-distillation. This yields compact, robust encodings—consistent with information bottleneck theory \cite{shwartz2017opening}—and reflects a more transparent optimization process for SAR recognition.

\section{Conclusion}
\label{sec:Con}
In this paper, we address the "representation trilemma" in CV-SAR recognition by proposing the Knowledge-Informed Neural Network. Our framework introduces a novel \textbf{"compression-aggregation-compression"} paradigm that synergistically integrates physical priors from the Electromagnetic Scattering Center model with semantic compression via self-distillation to learn compact and robust representations. Extensive experiments on five benchmarks confirm that KINN sets a new state-of-the-art in parameter-efficient recognition, outperforming existing methods, especially in limited-data and out-of-distribution scenarios. Thus, KINN serves as a strong case study for how a principled fusion of domain knowledge and deep learning can effectively address the representation trade-offs in a specialized scientific domain, with its core paradigm being readily extendable to other tasks and applications.

In the future study, we will extend the KINN concept to broader research fields, aiming to design low-parameter but well-generalized model by integrating different domain knowledge.


%

\ifCLASSOPTIONcompsoc

\ifCLASSOPTIONcaptionsoff
  \newpage
\fi

\bibliographystyle{IEEEtran}

\bibliography{main.bib}

\end{document}